\def\year{2020}\relax
\documentclass[letterpaper]{article} 
\usepackage{aaai20}  
\usepackage{times}  
\usepackage{helvet} 
\usepackage{courier}  
\usepackage[hyphens]{url}  
\usepackage{graphicx} 
\usepackage{graphicx}  
\usepackage[ruled,linesnumbered,noend]{algorithm2e}
\usepackage{amsfonts}       
\usepackage{amsmath}
\usepackage{xcolor}
\usepackage{tabulary}
\title{Spatiotemporally Constrained Action Space Attacks On Deep Reinforcement Learning Agents}
\author{Xian Yeow Lee,\textsuperscript{\rm 1} Sambit Ghadai,\textsuperscript{\rm 1} Kai Liang Tan,\textsuperscript{\rm 1} Chinmay Hegde,\textsuperscript{\rm 2} Soumik Sarkar\textsuperscript{\rm 1}\thanks{Corresponding Author}\\ 
\textsuperscript{\rm 1}Department of Mechanical Engineering, Iowa State University, Ames, IA 50011\\
\textsuperscript{\rm 2} Tandon School of  Engineering, New York University, Brooklyn, NY 11201\\
\{xylee, sambitg, kailiang, soumiks\}@iastate.edu,~chinmay.h@nyu.edu
}
\begin{document}

\maketitle

\begin{abstract}
Robustness of Deep Reinforcement Learning (DRL) algorithms towards adversarial attacks in real world applications such as those deployed in cyber-physical systems (CPS) are of increasing concern. Numerous studies have investigated the mechanisms of attacks on the RL agent's state space. Nonetheless, attacks on the RL agent's action space (corresponding to actuators in engineering systems) are equally perverse, but such attacks are relatively less studied in the ML literature. In this work, we first frame the problem as an optimization problem of minimizing the cumulative reward of an RL agent with decoupled constraints as the budget of attack. We propose the white-box Myopic Action Space (MAS) attack algorithm that distributes the attacks across the action space dimensions. Next, we reformulate the optimization problem above with the same objective function, but with a temporally coupled constraint on the attack budget to take into account the approximated dynamics of the agent. This leads to the white-box Look-ahead Action Space (LAS) attack algorithm that distributes the attacks across the action and temporal dimensions. Our results showed that using the same amount of resources, the LAS attack deteriorates the agent's performance significantly more than the MAS attack. This reveals the possibility that with limited resource, an adversary can utilize the agent's dynamics to malevolently craft attacks that causes the agent to fail. Additionally, we leverage these attack strategies as a possible tool to gain insights on the potential vulnerabilities of DRL agents.
\end{abstract}

\section{Introduction}

The spectrum of Reinforcement Learning (RL) applications ranges from engineering design and control~\cite{ed1,tan2019deep} to business~\cite{bus1} and creative design~\cite{art1}. As RL-based frameworks are increasingly deployed in real-world, it is imperative that the safety and reliability of these frameworks are well understood. While any adversarial infiltration of these systems can be costly, the safety of DRL systems deployed in cyber-physical systems (CPS) such as industrial robotic applications and self-driving vehicles are especially safety and life-critical. 

A root cause of these safety concerns is that in certain applications, the inputs to an RL system can be accessed and modified adversarially to cause the RL agent to take sub-optimal (or even harmful) actions. This is especially true when deep neural networks (DNNs) are used as key components (e.g., to represent policies) of RL agents. Recently, a wealth of results in the ML literature demonstrated that DNNs can be fooled to misclassify images by perturbing the input by an imperceptible amount~\cite{fgsm} or by introducing specific natural looking attributes~\cite{Joshi_2019_ICCV}. Such adversarial perturbations have also demonstrated the impacts of attacks on an RL agent's state space as shown by~\cite{huang2017adversarial}.

Besides perturbing the RL agent's state space, it is also important to consider adversarial attacks on the agent's \emph{action} space, which in engineering systems, represents physically manipulable actuators. We note that (model-based) actuator attacks have been studied in the cyber-physical security community, including vulnerability of continuous systems to discrete time attacks~\cite{AA1}; theoretical characteristics of undetectable actuator attacks~\cite{AA2}; and ``defense'' schemes that re-stabilizes a system when under actuation attacks~\cite{AA3}. However, the issue of adversarial attacks on a RL agent's action space has relatively been ignored in the DRL literature. In this work, we present a suite of novel attack strategies on a RL agent's action space. 

\noindent\textbf{Our contributions:} 
\begin{enumerate}
    \item  We formulate a white-box Myopic Action Space (MAS) attack strategy as an optimization problem with decoupled constraints.
    \item  We extend the formulation above by coupling constraints to compute a non-myopic attack that is derived from the agent's state-action dynamics and develop a white-box Look-ahead Action Space (LAS) attack strategy. Empirically, we show that LAS crafts a stronger attack than MAS using the same budget.
    \item  We illustrate how these attack strategies can be used to understand a RL agent's vulnerabilities. 
    \item  We present analysis to show that our proposed attack algorithms leveraging projected gradient descent on the surrogate reward function (represented by the trained RL agent model) converges to the same effect of applying projected gradient descent on the true reward function. 
\end{enumerate}

\begin{table*}[!t]
\caption{Landscape of adversarial attack strategies on RL agents. First column denotes if the attack takes into account the dynamics of the agent. Second column shows the method of computing the attacks; $O$ denotes an optimization-based method and $M$ denotes a model-based method where the parameters of a model needs to be learned. Last column represents if the attacks are mounted on agent's state space (S) or action space (A).}
\small
\centering
\begin{tabular}{l c c c}
\hline
\textbf{Method} & \textbf{Includes Dynamics} & \textbf{Method} & \textbf{Space of Attack} \\
\hline

FGSM on Policies~\cite{huang2017adversarial} & \text{\sffamily X} & O & S\\

ATN~\cite{tretschk2018sequential} & \text{\sffamily X} & M & S \\

Gradient based Adversarial Attack~\cite{pattanaik2018robust} & \text{\sffamily X} & O & S\\

Policy Induction Attacks~\cite{policy_induction} & \text{\sffamily X}  & O & S \\

Strategically-Timed and Enchanting Attack~\cite{lin2017tactics}  & \checkmark & O, M &  S \\

NR-MDP~\cite{tessler2019action} & \text{\sffamily X}  & M & A \\

\textbf{Myopic Action Space (MAS)}  & \text{\sffamily X}  & O & A\\

\textbf{Look-ahead Action Space (LAS)} & \checkmark  & O  & A  \\ 
\hline
\end{tabular}
\label{tab:method_compare}
\end{table*}

\section{Related Works}

Due to the large amount of recent progress in the area of adversarial machine learning, we only focus on reviewing the most recent attack and defense mechanisms proposed for DRL models. Table~\ref{tab:method_compare} presents the primary landscape of this area of research to contextualize our work.   

\subsection{Adversarial Attacks on RL Agent}

Several studies of adversarial attacks on DRL systems have been conducted recently.~\cite{huang2017adversarial} extended the idea of FGSM attacks in deep learning to RL agent's policies to degrade the performance of a trained RL agent. Furthermore,~\cite{policy_induction} showed that these attacks on the agent's state space are transferable to other agents. Additionally,~\cite{tretschk2018sequential} proposed attaching an Adversarial Transformer Network (ATN) to the RL agent to learn perturbations that will deceive the RL agent to pursue an adversarial reward. While the attack strategies mentioned above are effective, they do not consider the dynamics of the agent. One exception is the work by~\cite{lin2017tactics} that proposed two attack strategies. One strategy was to attack the agent when the difference in probability/value of the best and worst action crosses a certain threshold. The other strategy was to combine a video prediction model that predicts future states and a sampling-based action planning scheme to craft adversarial inputs to lead the agent to an adversarial goal, which might not be scalable.

Other studies of adversarial attacks on the specific application of DRL for path-finding have also been conducted by \cite{PF1} and \cite{PF2}, which results in the RL agent failing to find a path to the goal or planning a path that is more costly. 

\subsection{Robustification of RL Agents}

As successful attack strategies are being developed for RL models, various works on training RL agents to be robust against attacks have also been conducted.~\cite{pattanaik2018robust} proposed that a more severe attack can be engineered by increasing the probability of the worst action rather than decreasing the probability of the best action. They showed that the robustness of an RL agent can be improved by training the agent using these adversarial examples. More recently,~\cite{tessler2019action} presented a method to robustify RL agent's policy towards action space perturbations by formulating the problem as a zero-sum Markov game. In their formulation, a separate nominal and adversary policy are trained simultaneously with a critic network being updated over the mixture of both policies to improve both adversarial and nominal policies. Meanwhile,~\cite{havens2018online} proposed a method to detect and mitigate attacks by employing a hierarchical learning framework with multiple sub-policies. They showed that the framework reduces agent's bias to maintain high nominal rewards in the absence of adversaries. We note that other methods to defend against adversarial attacks exist, such as studies done by~\cite{tramer2017ensemble} and~\cite{sinha2018certifiable}. These works are done mainly in the context of a DNN but may be extendable to DRL agents that employs DNN as policies, however discussing these works in detail goes beyond the scope of this work.

\section{Mathematical Formulation}
\label{formulation}

\subsection{Preliminaries}

We focus exclusively on model-free RL approaches. Below, let $s_t$ and $a_t$ denote the (continuous, possibly high-dimensional) vector variables denoting \emph{state} and \emph{action}, respectively, at time $t$. We assume a state evolution function, $s_{t+1} = E(s_t, a_t)$ and let $R(s_t, a_t)$ denote the reward signal the agent receives for taking the action $a_t$, given $s_t$. The goal of the RL agent is to choose actions that maximizes the cumulative reward, $\sum_t R(s_t, a_t)$, given access to the trajectory, $\tau$, comprising all past states and actions. In value-based methods, the RL agent determines action at each time step by finding an intermediate quantity called the \emph{value function} that satisfies the recursive Bellman Equations. One example of such method is Q-learning~\cite{Watkins1992} where the agent discovers the Q-function, defined recursively as:
\[
Q_t(s_t,a_t) = R(s_t,a_t) + \max_{a'} Q_{t+1} (E(s_t,a_t),a') .
\]
The optimal action (or ``policy'') at each time step is to deterministically select the action that maximizes this Q-function conditioned on the observed state, i.e.,  
$$a^*_t = \arg \max_a Q(s_t,a).$$ 
In DRL, the Q-function in the above formulation is approximated via a parametric neural network $\Theta$; methods to train these networks include Deep Q-networks~\cite{mnih2015human}.

In policy-based methods such as policy gradients~\cite{sutton2000policy}, the RL agent \emph{directly} maps trajectories to policies. In contrast with Q-learning, the selected action is sampled from the policy parameterized by a probability distribution, $\pi = \mathbb{P}(a | s, \Theta)$, such that the expected rewards (with expectations taken over $\pi$) are maximized:
$$\pi^* = \arg \max_\pi E[R(\tau)],~~a_t^* \sim \pi^* .$$ 
In DRL, the optimal policy $\pi$ is the output of a parametric neural network $\Theta$, and actions at each time step are sampled; methods to train this neural network include proximal policy optimization (PPO)~\cite{schulman2017proximal}.

\subsection{Threat Model}

Our goal is to identify adversarial vulnerabilities in RL agents in a principled manner. To this end, We define a formal threat model, where we assume the adversary possesses the following capabilities: 
\begin{enumerate}
    \item \textbf{Access to RL agent's action stream}. The attacker can directly perturb the agent's nominal action adversarially (under reasonable bounds, elaborated below). The nominal agent is also assumed to be a closed-loop system and have no active defense mechanisms. 
    \item\textbf{ Access to RL agent's training environment}. This is required to perform forward simulations to design an optimal sequence of perturbations (elaborated below). 
    \item\textbf{Knowledge of trained RL agent's DNN}. This is needed to understand how the RL agent acts under nominal conditions, and to compute gradients. In the adversarial ML literature, this assumption is commonly made under the umbrella of {white-box attacks}.
\end{enumerate}

In the context of the above assumptions, the goal of the attacker is to choose a (bounded) action space perturbation that {minimizes} long-term discounted rewards. Based on how the attacker chooses to perturb actions, we define and construct two types of optimization-based attacks. We note that alternative approaches, such as training another RL agent to learn a sequence of attacks, is also plausible. However, an optimization-based approach is computationally more tractable to generate on-the-fly attacks for a target agent compared to training another RL agent (especially for high-dimensional continuous action spaces considered here) to generate attacks. Therefore, we restrict our focus on optimization-based approaches in this paper.

\subsection{Myopic Action-Space (MAS) Attack Model}

We first consider the case where the attacker is \emph{myopic}, i.e., at each time step, they design perturbations in a greedy manner without regards to future considerations. Formally, let $\delta_t$ be the action space perturbation (to be determined) and $b$ be a \emph{budget} constraint on the magnitude of each $\delta_t$ \footnote{Physically, the budget may reflect a real physical constraint, such as the energy requirements to influence an actuation, or it may be a reflection on the degree of imperceptibility of the attack.}. At each time step $t$, the attacker designs $\delta_t$ such that the anticipated future reward is minimized 
\begin{equation}
\begin{aligned}
& \underset{\delta_t}{\text{min}} \  R_\text{adv}(\delta_t) = R(s_t, a_t + \delta_t)+\sum_{j=t+1}^T R(s_j, a_j) \label{eq:mas} \\
& \text{subject to :} 
\ \|\delta_t\|_{p} \leq b, \\
& \ \ \ \ \ \ \ \ \ \ \ \ \ \ \ \ \ \  s_{j+1} = E(s_j, a_j), \\
& \ \ \ \ \ \ \ \ \ \ \ \ \ \ \ \ \ \ a_j = \Theta(s_j)~(\text{for}~j = t,\ldots,T),
\end{aligned}
\end{equation}
where $\|\cdot\|_p$ denotes the $\ell_p$-norm for some $p \geq 1$. Observe that while the attacker ostensibly solves separate (decoupled) problems at each time, the states themselves are not independent since given any trajectory, $s_{j+1} = E(s_j, a_j)$, where $E(s_j, a_j)$ is the transition of the environment based on $s_j$ and $a_j$. Therefore, ${\mathbf R}$ is implicitly coupled through time since it depends heavily on the evolution of state trajectories rather than individual state visitations. Hence, the adversary perturbations solved above are strictly myopic and we consider this a static attack on the agent's action space.

\subsection{Look-ahead Action Space (LAS) Attack Model}

Next, we consider the case where the attacker is able to look ahead and chooses a designed \emph{sequence} of future perturbations. Using the same notation as above, let $\sum_{j=t}^{t+H} R(s_j, a_j + \delta_j)$ denote the sum of rewards until a horizon parameter $H$, and let $\sum_{j = t + H + 1}^{T}R(s_j, a_j)$ be the future sum of rewards from time $j=t+H+1$. Additionally, we consider the (concatenated) matrix $\Delta = [\delta_{t}, \delta_{t+1} \dots \delta_{t+H}]$ and $B$ denote a budget parameter. The attacker solves the optimization problem:

\begin{equation}
\begin{aligned}
& \underset{\Delta}{\text{min}} \ R_{adv}(\Delta) =\sum_{j=t}^{t+H} R(s_j, a_j + \delta_j)+\sum_{j = t + H + 1}^{T} R(s_j, a_j) \label{eq:las} \\
& \text{subject to : } 
 \| \Delta \|_{p, q} \leq B, \Delta = [\delta_t, \delta_{t+1}, \dots, \delta_H], \\
& \ \ \ \ \ \ \ \ \ \ \ \ \ \ \ \ \ \ s_{j+1} = E(s_j, a_j), \\
& \ \ \ \ \ \ \ \ \ \ \ \ \ \ \ \ \ \ a_j = \Theta(s_j)  
\end{aligned}
\end{equation}
where $\|\cdot\|_{p,q}$ denotes the  $\ell_{p,q}$-norm \cite{boyd2004convex}. By coupling the objective function and constraints through the temporal dimension, the solution to the optimization problem above is then forced to take the dynamics of the agent into account in an explicit manner. 

\section{Proposed Algorithms}
In this section, we present two attack algorithms based on the optimization formulations presented in previous section.

\subsection{Algorithm for Mounting MAS Attacks}

We observe that \eqref{eq:mas} is a nonlinear constrained optimization problem; therefore, an immediate approach to solve it is via projected gradient descent (PGD). Specifically, let $\mathcal{S}$ denote the $\ell_p$ ball of radius $b$ in the action space. We compute the gradient of the adversarial reward, $\nabla R_\text{adv}$ w.r.t. the action space variables and obtain the \emph{unconstrained} adversarial action $\hat{a}_{t + \frac{1}{2}}$ using gradient descent with step size $\eta$. Next, we calculate the \emph{unconstrained} perturbation $\delta_t$ and project in onto $\mathcal{S}$ to get $\delta'_t$ :

\begin{equation}
\begin{aligned}
& \hat a_{t + \frac{1}{2}} = a_t  - \eta \nabla R_{adv}(s_t, \hat a_t), 
\\ &\delta_t = \hat{a}_{t + \frac{1}{2}} - a_t,
\\ &\delta'_t = P_{\mathcal{S}}(\delta_{t}).
\end{aligned}
\end{equation}

Here, $a_t$ represents the nominal action. We note that this approach resembles the fast gradient-sign method (FGSM)~\cite{fgsm}, although we compute standard gradients here. As a variation, we can compute multiple steps of gradient descent w.r.t the action variable prior to projection; this is analogous to the basic iterative method (or iterative FGSM)~\cite{kurakin2016adversarial}. The MAS attack algorithm is shown in the supplementary material. 

We note that in DRL approaches, only a \emph{noisy proxy} of the true reward function is available: In value-based methods, we utilize the learned Q-function (for example, a DQN) as an approximate of the true reward function, while in policy-iteration methods, we use the probability density function returned by the optimal policy as a proxy of the reward, under the assumption that actions with high probability induces a high expected reward. Since DQN selects
the action based on the argmax of Q-values and policy iteration samples the action with highest probability, the Q-values/action-probability remains a useful proxy for the reward in our attack formulation. Therefore, our proposed MAS attack is technically a version of \emph{noisy projected gradient descent} on the policy evaluation of the nominal agent. We elaborate on this further in the theoretical analysis section.

\subsection{Algorithm for Mounting LAS Attacks}

\begin{algorithm}[!t]
\SetAlgoLined
\SetNoFillComment
\DontPrintSemicolon
\caption{Look-ahead Action Space (LAS) Attack}
\label{alg:projected}
Initialize nominal and adversary environments $E_{nom}$, $E_{adv}$ with same random seed\\
Initialize nominal agent $\pi_{nom}$ weights, $\theta$ \\
Initialize budget $B$, adversary action buffer $A_{adv}$, horizon $H$\\
\While {$t \leq T$}
    {
    Reset $A_{adv}$ \\
    \uIf{$H$ = 0}{
        Reset $H$ and $B$}
    \While {$k \leq H$}
        {
        Compute gradient of surrogate reward $\nabla R_{adv}$ \\
        Compute adversarial action $\hat a_{t + \frac{1}{2},k}$ using $\nabla R_{adv}$ \\ 
        Compute $\delta_{t,k}= \hat a_{t + \frac{1}{2},k} - a_{t,k}$\\
        Append $\delta_{t,k}$ to $A_{adv}$\\
        Step through $E_{adv}$ with $a_{t,k}$ to get next state
        }
    Compute $||\delta_{t,k}||_{\ell_p}$ for each element in $A_{adv}$ \\
    
    Project sequence of $||\delta_{t,k}||_{\ell_p}$ in $A_{adv}$ on to ball of size $B$ to obtain look-ahead sequence of budgets [$b_{t,k}$, $b_{t, k+1}$ $\dots$ $b_{t, k+H}$]\\
    
    Project each $\delta_{t, k}$ in $A_{adv}$ on to look-ahead sequence of budgets computed in the previous step to get sequence [$\delta'_{t,k}$ $\delta'_{t,k+1}$ $\dots$ $\delta'_{t,k+H}$]\\
    
    Compute projected adversarial action  $\hat a_{t}$ = $a_{t} + \delta'_{t,k}$ \\
    
    Step through $E_{nom}$ with $\hat a_{t}$\\
    $B \leftarrow max(0, B - \delta'_{t,k})$\\
    $H \leftarrow H-1$}
\end{algorithm}

The previous algorithm is myopic and can be interpreted as a purely \emph{spatial} attack. In this section, we propose a \emph{spatiotemporal} attack algorithm by solving Eq.~\eqref{eq:las} over a given time window $H$. Due to the coupling of constraints in time, this approach is more involved. We initialize a copy of both the nominal agent and environment, called the adversary and adversarial environment respectively. At time $t$, we sample a virtual roll-out trajectory up until a certain horizon $t + H$ using the pair of adversarial agent and environment. At each time step $k$ of the virtual roll-out, we compute action space perturbations $\delta_{t,k}$ by taking (possibly multiple) gradient updates. Next, we compute the norms of each $\delta_{t,k}$ and project the sequence of norms back onto an $\ell_q$-ball of radius $B$. The resulting projected norms at each time point now represents the individual budgets, $b_k$, of the spatial dimension at each time step. Finally, we project the original $\delta_{t,k}$ obtained in the previous step onto the $\ell_p$-balls of radii $b_k$, respectively to get the final perturbations $\delta'_{t,k}$.\footnote{Intuitively, these steps represent the allocation of overall budget $B$ across different time steps.} We note that to perform virtual roll-outs at every time step $t$, the state of the $E_{{adv}}$ has to be the same as the state of $E_{nom}$ at $t$. To accomplish this, we saved the history of all previous actions to re-compute the state of the $E_{adv}$ at time $t$ from $t=0$. While this current implementation may be time-consuming, we believe that this problem can be avoided by giving the adversary direct access to the current state of the nominal agent through platform API-level modifications; or explicit observations (in real-life problems).

In subsequent time steps, the procedure above is repeated with a reduced budget of $B - \sum^t_{t=0}\delta'_t$ and reduced horizon $H-t$ until $H$ reaches zero. The horizon $H$ is then reset again for planning a new spatiotemporal attack. An alternative formulation could also be shifting the window without reducing its length until the adversary decides to stop the attack. However, we consider the first formulation such that we can compare the performance of LAS with MAS for an equal overall budget. This technique of re-planning the $\delta'_t$ at every step while shifting the window of $H$ is similar to the concept of \emph{receding horizons} regularly used in optimal control~\cite{RC1}. It is evident that using this form of dynamic re-planning mitigates the planning error that occurs when the actual and simulated state trajectories diverge due to error accumulation~\cite{MPC}. Hence, we perform this re-planning at every $t$ to account for this deviation. The pseudocode is provided in Alg.~\ref{alg:projected}.

\begin{figure}[!t]
    \centering
    \includegraphics[width=0.8\columnwidth]{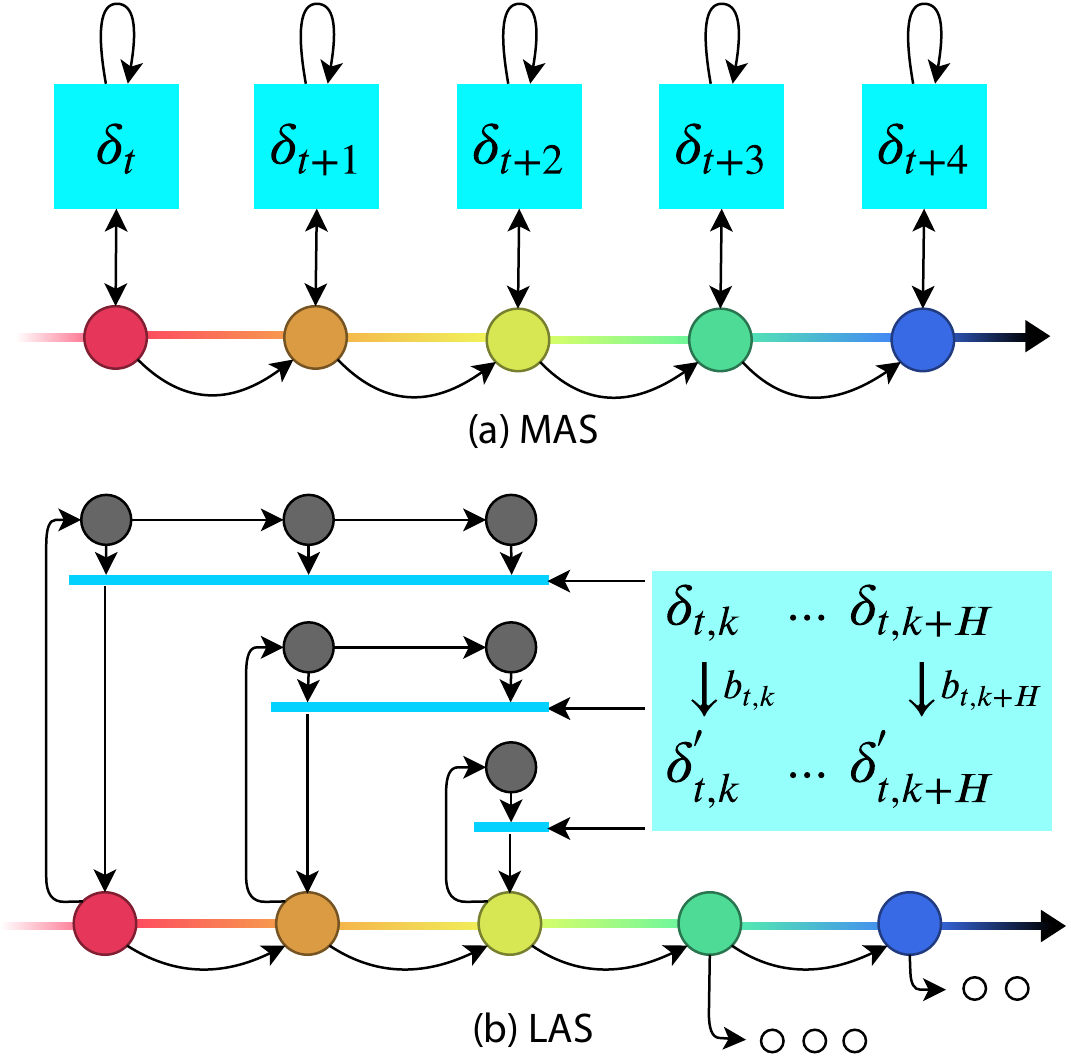}
    \caption{Visual comparison of MAS and LAS. In MAS, each $\delta_t$ is computed via multi-step gradient descent w.r.t. expected rewards for the current step. In LAS, each $\delta_{t,k}^{'}$ is computed w.r.t. the dynamics of the agent with receding horizon. An adversarial agent \& environment is used to compute LAS for each step. Projection is applied to each $\delta_t$ in the temporal domain. The final perturbed action is obtained by adding the first $\delta_{t,k}^{'}$ to the nominal action. This is done until the end of the attack window, i.e., $H-t=0$.
    }
    \label{fig:framework}
\end{figure}

\subsection{Theoretical Analysis}
We can show that projected gradient descent on the surrogate reward function (modeled by the RL agent network) to generate both MAS and LAS attacks provably converges; this can be accomplished since gradient descent on a surrogate function is akin to a noisy gradient descent on the true adversarial reward. 

As described in previous sections, our MAS/LAS algorithms are motivated in terms of the adversarial reward $R_{adv}$. However, if we use either DQN or policy gradient networks, we do not have direct access to the reward function, but only its noisy \emph{proxy}, defined via a neural network. Therefore, we need to argue that performing (projected) gradient descent using this proxy loss function is a sound procedure. To do this, we appeal to a recent result by~\cite{ge2015escaping}, who prove convergence of noisy gradient descent approximately converges to a local minimum. More precisely, consider a general constrained nonlinear optimization problem:
\begin{align*}
    &\min f(x) \\ 
    & \text{s.t.}~c(x) = 0,
\end{align*}
where $c$ is an arbitrary (differentiable, possibly vector-valued) function encoding the constraints. Define $S = \{x | c(x) = 0 \}$ define the constraint set. We attempt to minimize the objective function via noisy  (projected) gradient updates:
\begin{align*}
    x_{t + 1/2} &= x_t - \eta \nabla f(x_t) + \xi_t \, , \\
    x_{t+1} &= P_S(x_{t+1/2}) .
\end{align*}

\noindent\textbf{Theorem 1.} \textit{(Convergence of noisy projected gradients.)} 
\emph{Assume that the noise terms $\{\xi_t\}$ are i.i.d., satisfying $E[\xi] = 0, E[\xi \xi^T] = \sigma^2 Id$, $\|\xi\| \leq O(1)$ almost surely. Assume that both the constraint function $c(\dot)$ and the objective function $f(\cdot)$ is $\beta$-smooth, $L$-Lipschitz, and possesses $\rho_i$-Lipschitz Hessian. Assume further that the objective function $f$ is $B$-bounded. Then, there exists a learning rate $\eta = O(1)$ such that with high probability, in $\text{polylog}(1/\eta^2)$ iterations, noisy projected gradient descent converges to a point $\hat{x}$ that is $\text{polylog}(\sqrt{\eta})$-close to some local minimum of $f$.}

In our case, $f$ and $\xi$ depends on the structure of the RL agent's neural network. (Smoothness assumptions of $f$ can perhaps be justified by limiting the architecture of the network, but the iid-ness assumption on $\xi$ is hard to verify). As such, it is difficult to ascertain whether the assumptions of the above theorem are satisfied in specific cases. Nonetheless, an interesting future theoretical direction is to understand Lipschitz-ness properties of specific families of DQN/policy gradient agents. 

We defer further analysis of the double projection step onto mixed-norm balls used in our proposed LAS algorithms to the supplementary material.


\section{Experimental Results \& Discussion}

To demonstrate the effectiveness and versatility of our methods, we implemented them on RL agents with continuous action environments from OpenAI's gym~\cite{gym} as they reflect the type of action space in most practical applications~\footnote{Codes and links to supplementary are available at \url{https://github.com/xylee95/Spatiotemporal-Attack-On-Deep-RL-Agents}}. For policy-based methods, we trained a nominal agent using the PPO algorithm and a DoubleDQN (DDQN) agent~\cite{ddqn} for value-based methods\footnote{The only difference in implementation of policy vs value-based methods is that in policy methods, we take analytical gradients of a distribution to compute the attacks (e.g., in line 10 of Algorithm~\ref{alg:projected}) while for value-based methods, we randomly sample adversarial actions to compute numerical gradients.}. Additionally, we utilize Normalized Advantage Functions~\cite{gu2016continuous} to convert the discrete nature of DDQN's output to continuous action space. For succinctness, we present the results of the attack strategies only on PPO agent for the Lunar-Lander environment. Additional results of DDQN agent in Lunar Lander and Bipedal-Walker environments and PPO agent in Bipedal-Walker, Mujoco Hopper, Half-Cheetah and Walker environments are provided in the supplementary materials. As a baseline, we implemented a random action space attack, where a random perturbation bounded by the same budget $b$ is applied to the agent's action space at every step. For MAS attacks, we implemented two different spatial projection schemes, $\ell_1$ projection based on \cite{l1_proj} that represents a sparser distribution and $\ell_2$ projection that represents a denser distribution of attacks. For LAS attacks, all combinations of spatial and temporal projection for $\ell_1$ and $\ell_2$ were implemented.

\begin{figure*}[!ht]
    \centering
    \includegraphics[width=0.8\textwidth]{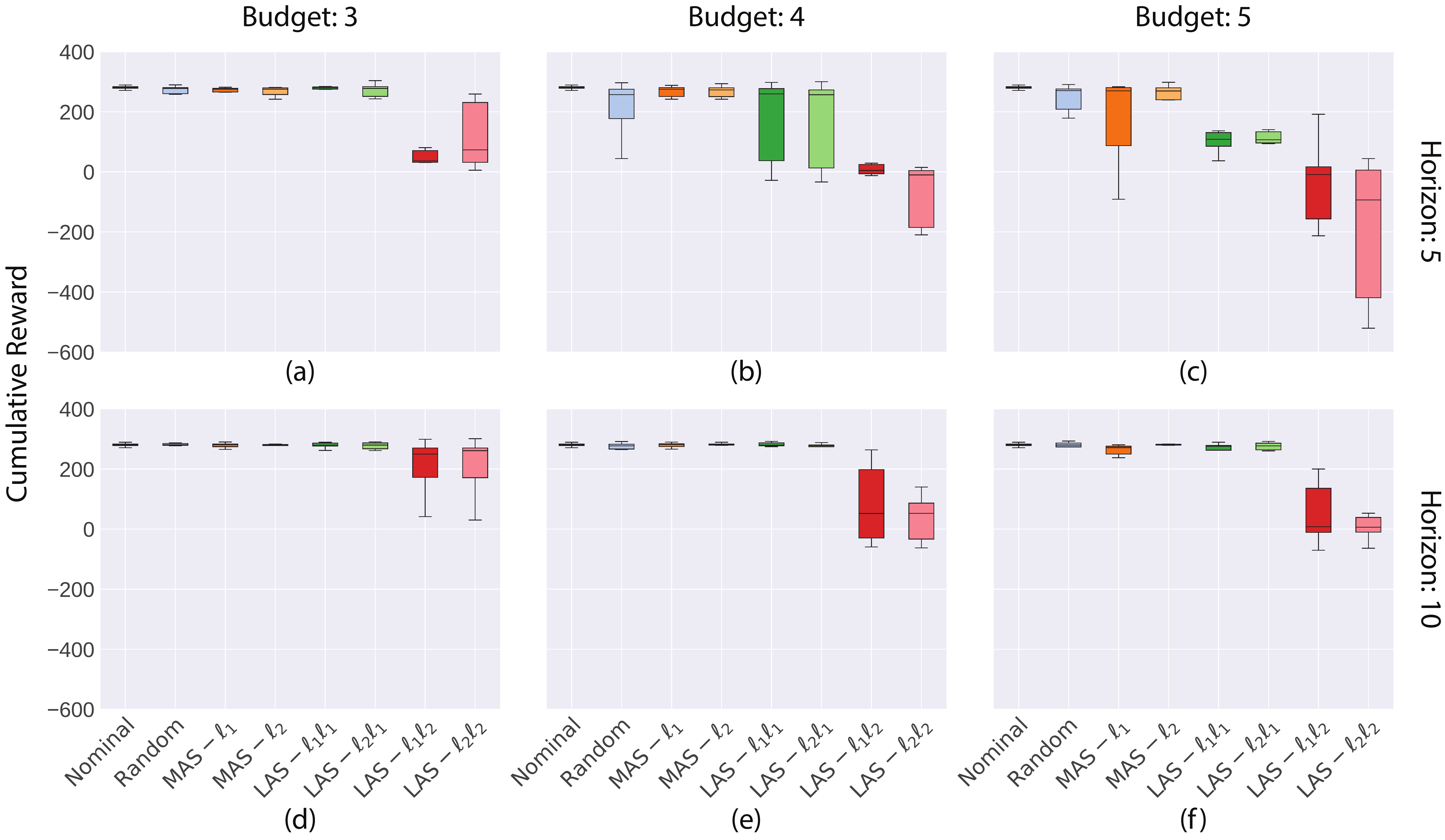}
    \caption{Box plots of PPO Lunar Lander showing average cumulative reward across 10 episodes for each attack methods. The top Figs. (a-c) have $H$=5 with $B$= 3, 4, and 5 respectively. For a direct comparison, corresponding MAS budgets are taken as $b=B/H$. Similarly, Figs.(d-f) have the same $B$ values but with $H$=10. An obvious trend is that as $B$ increases, the effectiveness of LAS over MAS becomes more evident as seen in the decreasing trend of the reward.}
    \label{fig:PPO_LL_boxplot}
\end{figure*}

\subsection{Comparison of MAS and LAS Attacks}

Fig.~\ref{fig:PPO_LL_boxplot} shows distributions of cumulative rewards obtained by the PPO agent across ten episodes in a Lunar Lander environment, with each subplot representing different combinations of budget, $B$ and horizon, $H$. Top three subplots show experiments with a $H$ value of 5 time steps and $b$ value of 3, 4, and 5 from left to right respectively. Bottom row of figures show a similar set of experiments but with a $H$ value of 10. For a direct comparison between MAS and LAS attacks with equivalent budgets across time, we have assigned the corresponding MAS budget values as $b = B/H$. This assumes that the total budget $B$ is allocated uniformly across every time step for a given $H$, while LAS has the flexibility to allocate the attack budget non-uniformly in the same interval, conditioned on the dynamics of the agent.

We note that keeping $H$ constant while increasing $B$ provides both MAS and LAS with a higher budget to inject $\delta_t$ to the nominal actions. We observe that with a low budget of $3$ (Fig.~\ref{fig:PPO_LL_boxplot}a), only LAS is successful in attacking the RL agent, as seen by the corresponding decrease in rewards. With a higher budget of 5 (Fig.~\ref{fig:PPO_LL_boxplot}c), MAS has a more apparent effect on the performance of the RL agent while LAS reduces the performance of the agent severely.

With $B$ constant, increasing $H$ allows the allocated $B$ to be distributed along the increased time horizon. In other words, LAS virtually looks-ahead further into the future. In the most naive case, a longer horizon dilutes the severity of each $\delta_t$ in compared to shorter horizons. By comparing similar budget values of different horizons (i.e horizons 5 and 10 for budget 3, Fig.~\ref{fig:PPO_LL_boxplot}a and Fig.~\ref{fig:PPO_LL_boxplot}d respectively), attacks for $H=10$ are generally less severe than their $H=5$ counterparts. For all $B$ and $H$ combinations, we observe that MAS attacks are generally less effective compared to LAS. We note that this is a critical result of the study as most literature on static attacks have shown that the attacks can be ineffective below a certain budget. Here, we demonstrate that while MAS attacks can seemingly look ineffective for a given budget, a stronger and more effective attack can essentially be crafted using LAS with the same budget. 

In the following sections, we further study the difference between MAS and LAS as well as demonstrate how the attacks can be utilized to understand the vulnerabilities of the agent in different environments. 

\begin{figure*}[!t]
    \centering
    \includegraphics[width=0.99\textwidth]{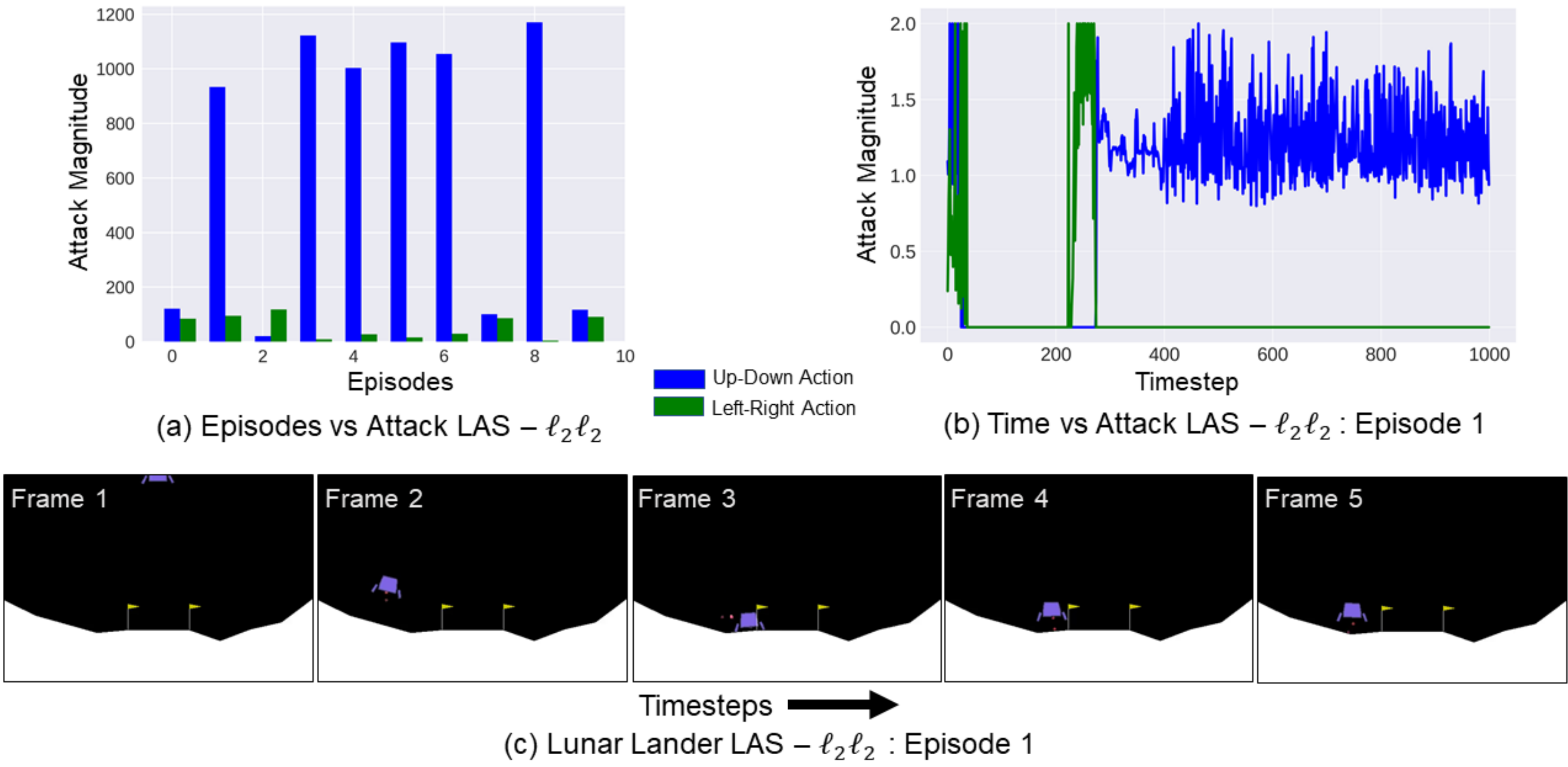}
    \caption{Time vs Attack magnitude along action dimension for LAS attacks with $B = 4, H=5$ in Lunar Lander environment with PPO RL agent. (a) Variation of attack magnitude along Up-Down and Left-Right action dimensions through different episodes. In all episodes except episode 2, Up-Down action is more heavily attacked than Left-Right. (b) Variation of attack magnitude through time for episode 1 of (a). After 270 steps, the agent is not attacked in the Left-Right dimension, but heavily attacked in Up-Down directions. (c) Actual rendering of Lunar Lander environment for episode 1 of (a) corresponding to (b). Frame 1-5 are strictly increasing time steps showing trajectory of the RL agent controlling the lunar lander.}
    \label{fig:timedelta_ppo}
\end{figure*}

\subsection{Action Dimension Decomposition of LAS Attacks}

Fig. \ref{fig:timedelta_ppo} shows action dimension decomposition of LAS attacks. The example shown in Fig. \ref{fig:timedelta_ppo} is the result of $\ell_2$ projection in action space with $\ell_2$ projection in time. From Fig. \ref{fig:timedelta_ppo}a, we observe that through all the episodes of LAS attacks, one of the action dimension (i.e., Up - Down direction of lunar lander) is consistently perturbed more, i.e., accumulates more attack, than Left-Right direction.

Fig. \ref{fig:timedelta_ppo}b shows a detailed view of action dimension attacks for an episode (Episode 1). It is evident from the figure that the Up-Down actions of the lunar lander are more prone to attacks throughout the episode than Left-Right actions. Additionally, Left-Right action attacks are restricted after certain time steps and only the Up-Down actions are attacked further. Fig. \ref{fig:timedelta_ppo}c further corroborates the observation in the Lunar Lander environment. As the episode progresses in Fig. \ref{fig:timedelta_ppo}c, the lunar lander initially lands on the ground in frame 3, but lifts up and hovers until the episode ends in frame 5. This observation supports the fact that the proposed attacks are effective in perturbing the action dimensions in an optimal manner; as in this case, perturbing the lunar lander in the horizontal direction will not further decrease rewards. On the other hand, hovering the lunar lander will cause the agent to use more fuel, which consequently decreases the total reward. From these studies, it can be concluded that LAS attacks (correlated with projections of actions in time) can clearly isolate vulnerable action dimension(s) of the RL agent to mount a successful attack. 

\subsection{Ablation Study of Horizon and Budget}

Lastly, we performed multiple ablation studies to compare the effectiveness of LAS and MAS attacks. While we have observed that LAS attacks are generally stronger than MAS, we hypothesize that there will be an upper limit to LAS's advantage as the allowable budget increases. We take the difference of each attack's reduction in rewards (i.e. attack - nominal) and visualize how much rewards LAS reduces as compared to MAS under different conditions of $B$ and $H$. In the case of PPO in Lunar Lander, we observe that the reduction in rewards of LAS vs MAS becomes less drastic as budget increases, hence showing that LAS has diminishing returns as both MAS and LAS saturates at higher budgets. We defer detailed discussions and additional figures of the ablation study to the supplementary materials.

\section{Conclusion \& Future Work}

In this study, we present two novel attack strategies on an RL agent's action space; a myopic attack (MAS) and a non-myopic attack (LAS). The results show that LAS attacks, that were crafted with explicit use of the agent's dynamics information, are more powerful than MAS attacks. Additionally, we observed that applying LAS attacks on RL agents reveals the possible vulnerable actuators of an agent, as seen by the non-uniform distribution of attacks on certain action dimensions. This can be leveraged as a tool to identify the vulnerabilities and plan a mitigation strategy under similar attacks. Possible future works include extending the concept of LAS attacks to state space attacks where the agent's observations are perturbed instead of the agent's actions while taking into account the dynamics of the agent. Additionally, while we did not focus on the imperceptibility and deployment aspects of the proposed attacks in this study, defining a proper metric in terms of detectability in action space and optimizing the budget to remain undetected for different environments will be a future research direction. 

\section{Acknowledgement}
This work was supported in part by NSF grants CNS-1845969 and CCF-2005804, and AFOSR YIP Grant FA9550-17-1-0220.

\fontsize{10.0pt}{10.0pt}\selectfont
\bibliography{ref}
\bibliographystyle{aaai}
\fontsize{10.0pt}{10.0pt}\selectfont
\clearpage
\newpage


\appendix

\section{Pseudocode of MAS Attack}
\begin{algorithm}[ht]
\SetAlgoLined
\SetNoFillComment
\DontPrintSemicolon
\caption{Myopic Action Space (MAS) Attack}
\label{alg:static}
Initialize nominal environment, $E_{nom}$, nominal agent $\pi_{nom}$ with weights, $\theta$ \\
Initialize budget $b$\\
\While {$t$ $\leq$ $T$}{
    Compute gradient of surrogate reward $\nabla R_{adv}$ \\
    Compute adversarial action $\hat a_{t + \frac{1}{2}}$ using $\nabla$ $R_{adv}$ \\
    Compute $\delta_t= \hat a_{t + \frac{1}{2}} - a_{t}$, project $\delta_t$ onto ball of size $b$ to get $\delta'_t$\\
    Compute projected adversarial action  $\hat a_{t}$ = $a_{t} + \delta'_t$ \\
    Step through $E_{nom}$ with $\hat a_{t}$ to get next state}
\end{algorithm}

\section{Analysis}
\subsection{Projections onto Mixed-norm Balls}
\label{sec:proj}

The Look-ahead Action Space (LAS) Attack Model described above requires projecting onto the mixed-norm $\ell_{p,q}$-ball of radius $B$ in a vector space. Below, we show how to provably compute such projections in a computationally efficient manner. For a more complete treatment, we refer to \cite{sra2012fast}. Recall the definition of the $(p,q)$-norm. Let $X \in \mathbb{R}^{m \times n}$ be partitioned into sub-vectors $x_i,~i \in [n]$ of length-$m$. Then,
$$
\|X\|_{p,q} := \left(\sum_{i=1}^n \|x_i\|_q^p \right)^{1/p} .
$$
Due to scale invariance of norms, we can assume $B = 1$. We consider the following special cases:

\begin{enumerate}
    \item $p=1, q=1$: this reduces to the case of the ordinary $\ell_1$-norm in $\mathbb{R}^{mn}$. Projection onto the unit $\ell_1$-ball can be achieved via \emph{soft-thresholding} every entry in $X$:
    $$
    P_S(X_{i,j}) = \text{sign}(X_{i,j}) \cdot (|X_{i,j}| - \lambda)_+ , 
    $$
    where $\lambda > 0$ is a KKT parameter that can be discovered by a simple sorting the (absolute) values of $X$. See~\cite{condat2016fast}.
    \item $p = 2, q = 2$: this reduces to the case of the isotropic $\ell_2$-norm in $\mathbb{R}^{mn}$. Projection onto the unit $\ell_2$-ball can be achieved by simple normalization:
    $$
    P_S(X_{i,j}) = X_{i,j} / \|X\|_{2,2} .
    $$
    \item $p = 1, q = 2$: this is a ``hybrid'' combination of the above two cases, and corresponds to the procedure that we use in mounting our LAS attack. Projection onto this ball can be achieved by a three-step method. First, we compute the $n$-dimensional vector, $v$, of column-wise $\ell_2$-norms. Then, we project $v$ onto the unit $\ell_1$-ball; essentially, this enables us to ``distribute'' the (unit) budget across columns. Since $\ell_1$-projection is achieved via soft-thresholding, a number of coordinates of this vector are zeroed out, and others undergo a shrinkage. Call this (sparsified) projected vector $v_p$. Finally, we perform an $\ell_2$ projection, i.e., we scale each column of $X$ by dividing by its norm and multiplying by the entries of $v_p$:
    $$
    P_S(X_{i,j}) = \frac{X_{i,j}}{\|x_j\|_2} \cdot v_p(i) .
    $$
\end{enumerate}

\section{Additional Experiments}

\subsection{Comparison of Attacks Mounted on PPO Agent in Bipedal-Walker Environment}

\begin{figure*}[ht]
    \centering
    \includegraphics[width=0.8\textwidth,clip,trim={0.0in 0.0in 0.0in 0.0in}]{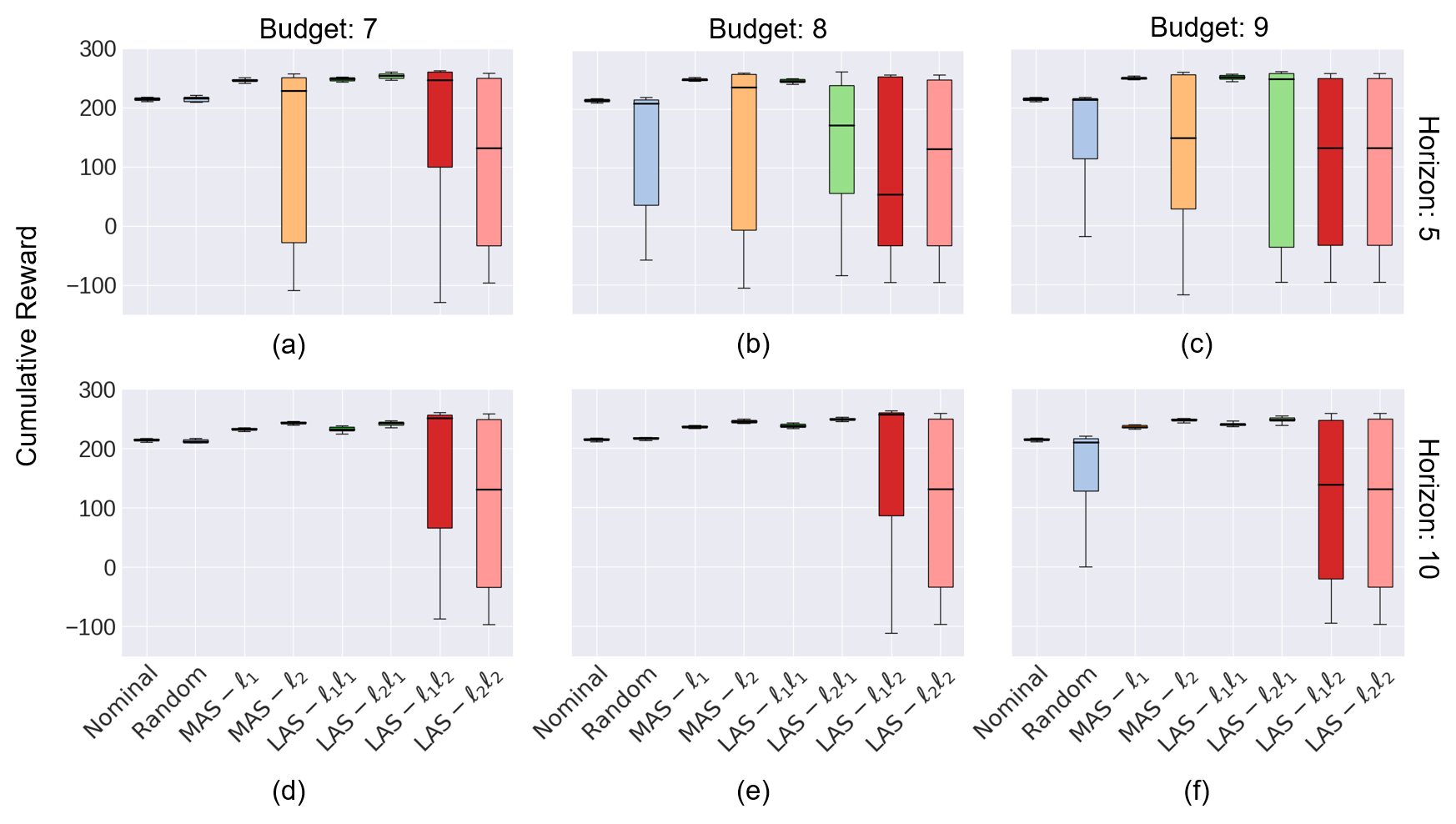}
    \caption{Boxplots showing cumulative rewards of PPO agent in Bipedal-Walker environment under different attack strategies across 10 different episodes. Plots (a), (b) and (c) are attacked with a horizon of 5 time steps with budget value of 7, 8, and 9 respectively. (d), (e), and (f) are attacked with a horizon value of 10 time steps with budget value $B$ of 7, 8, and 9. }
    \label{fig:PPO_BW_boxplot}
\end{figure*}

The results in Fig.~\ref{fig:PPO_BW_boxplot} depicts the comparison between the MAS and LAS attacks applied on a PPO agent in the Bipedal-Walker environment. A similar trend is observed where LAS attacks are generally more severe than MAS attacks. We acknowledge that in this environment, MAS attacks are sometimes effective in reducing the rewards as well. However, this can be attributed to the Bipedal Walker having more dimensions (4 dimensions) in terms of it's action space in compared to the Lunar-Lander (2 dimensions) environment. In addition, the actions of the Bipedal Walker is also highly coupled, in compared to the actions of the Lunar Lander. Hence, the agent for Bipedal-Walker is more sensitive towards perturbations, which explains the increase efficacy of MAS attacks.

\subsection{Comparison of Attacks Mounted on DDQN Agent in Lunar Lander and Bipedal-Walker Environments}

\begin{figure*}[ht]
    \centering
    \includegraphics[width=0.85\textwidth,clip,trim={0.0in 0.0in 0.0in 0.0in}]{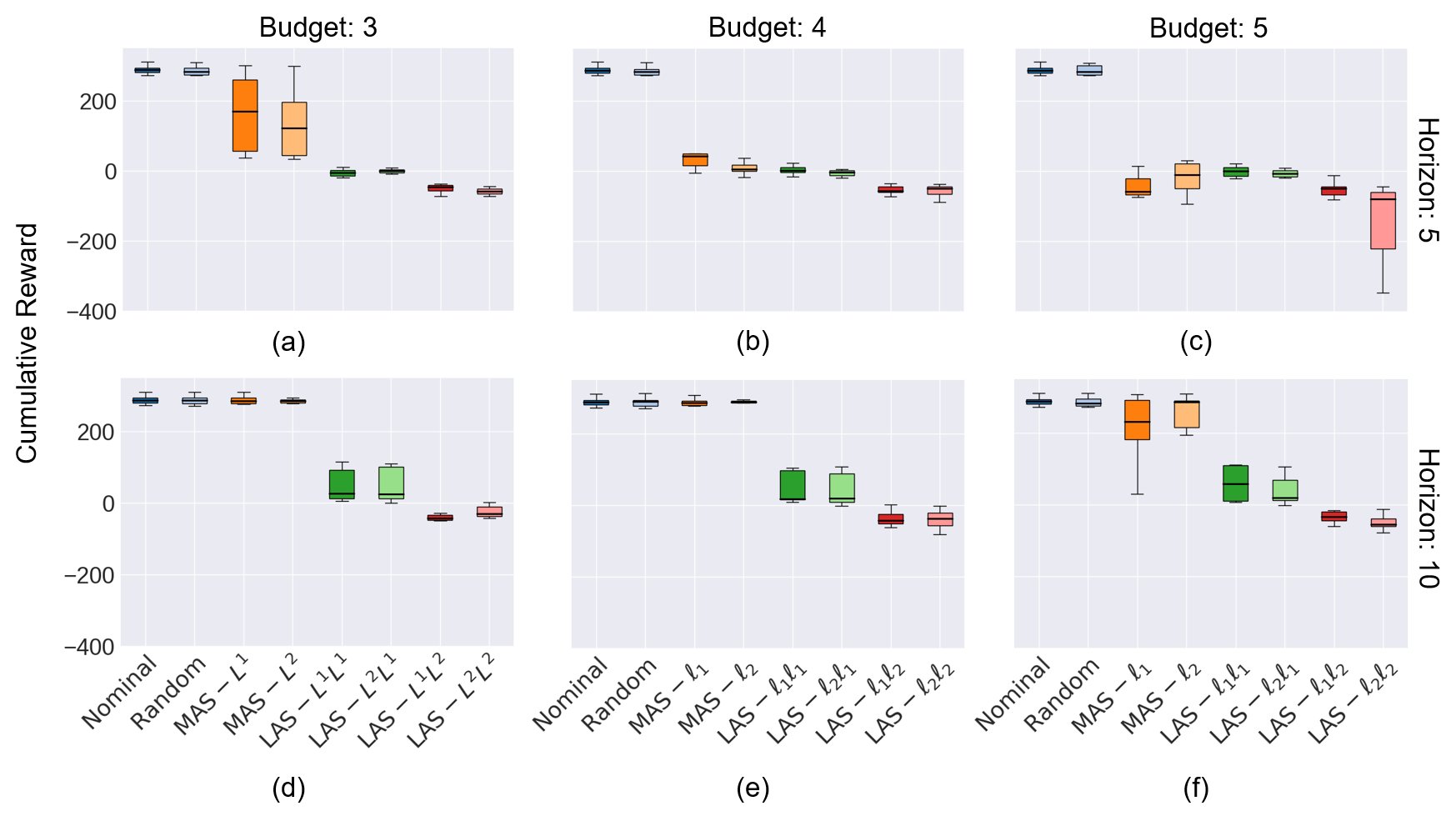}
    \caption{
    DDQN Lunar Lander box plots showing average cumulative reward across 10 episodes for each attack method. Plots (a), (b) and (c) are attacked with a horizon of 5 time steps with budget value of 3, 4, and 5 respectively. (d), (e), and (f) are attacked with a horizon value of 10 time steps with budget value of 3, 4, and 5 respectively. Given the same horizon and budget, it is evident LAS attacks are more severe than MAS attacks, which in turn are generally more effective than random attacks.}
    \label{fig:DDQN_LL_boxplot}
\end{figure*}

\begin{figure*}[h]
    \centering
    \includegraphics[width=0.85\textwidth,clip,trim={0.0in 0.0in 0.0in 0.0in}]{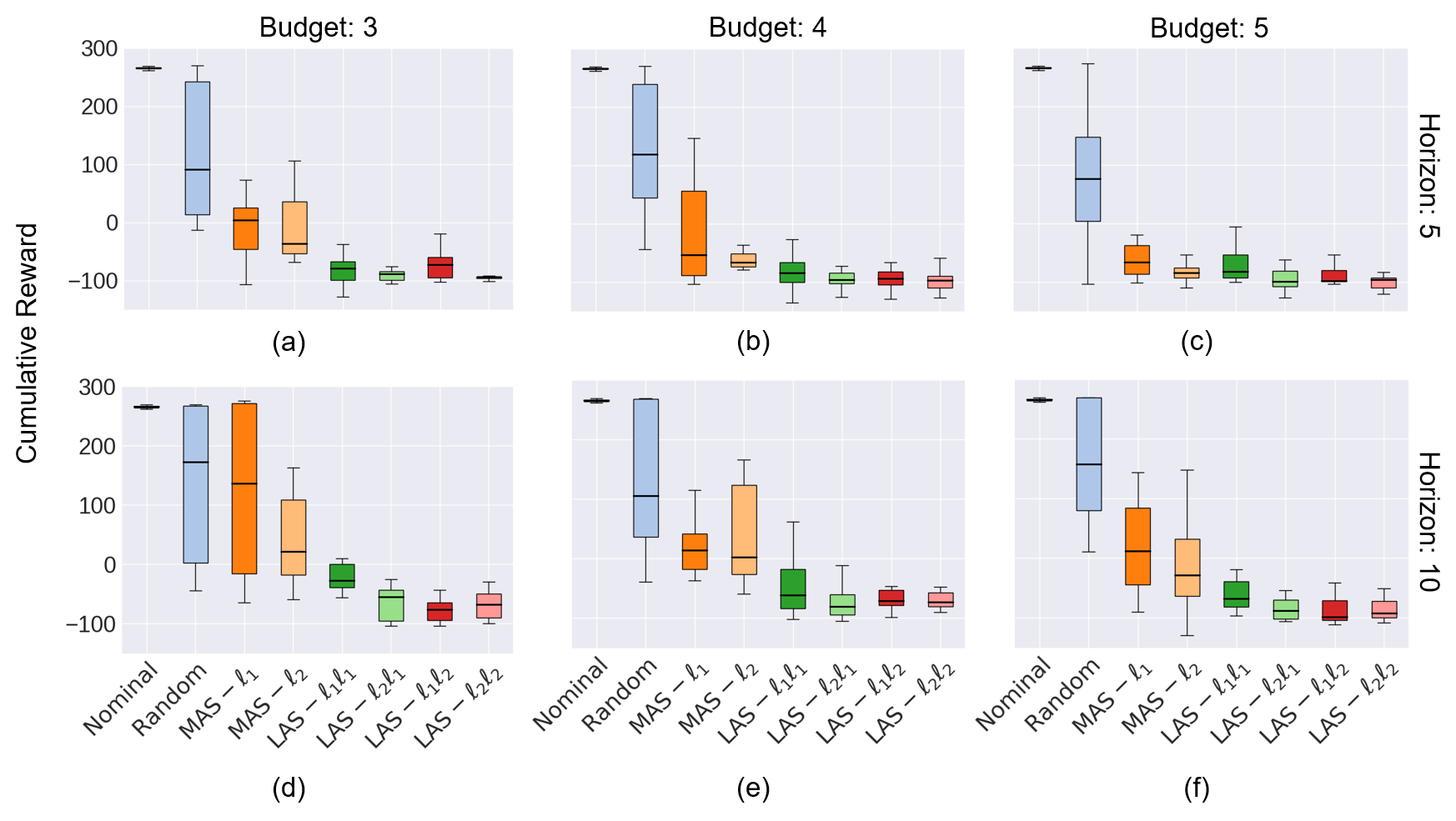}
    \caption{DDQN Bipedal Walker box plots showing average cumulative reward across 10 episodes for each attack method. Plots (a), (b) and (c) are attacked with a horizon of 5 time steps with budget value of 3, 4, and 5 respectively. (d), (e), and (f) are attacked with a horizon value of 10 time steps with budget value of 3, 4, and 5 respectively. Given the same horizon and budget, it is evident LAS attacks are more severe than MAS attacks, which in turn are generally more effective than random attacks.} 
    \label{fig:DDQN_BW_boxplot}
\end{figure*}

In Figures~\ref{fig:DDQN_LL_boxplot} and~\ref{fig:DDQN_BW_boxplot}, we present additional results on the efficacy of different attack strategies for a DoubleDQN agent trained in the Lunar Lander and Bipedal Walker environment. An interesting observation is that in both environment, the effects of the attacks are more severe for the DDQN agent in compared to the PPO agent as seen in the box-plots. We conjecture that this is due to the deterministic nature of DDQN method where the optimal action is unique. In contrast, the PPO agent was trained with a certain stochastic in the process, which might explain the additional robustness of PPO agent to noise or perturbations. Nevertheless, in the DDQN agent, we still observe a similar trend where given the same about of budget, more severe attacks can be crafted using the LAS strategy in compared to MAS or random attacks.

\subsection{Comparison of Attacks Mounted on PPO Agent Mujoco Control Environments}

\begin{figure*}[!ht]
    \centering
    \includegraphics[width=0.85\textwidth,clip,trim={0.0in 0.0in 0.0in 0.0in}]{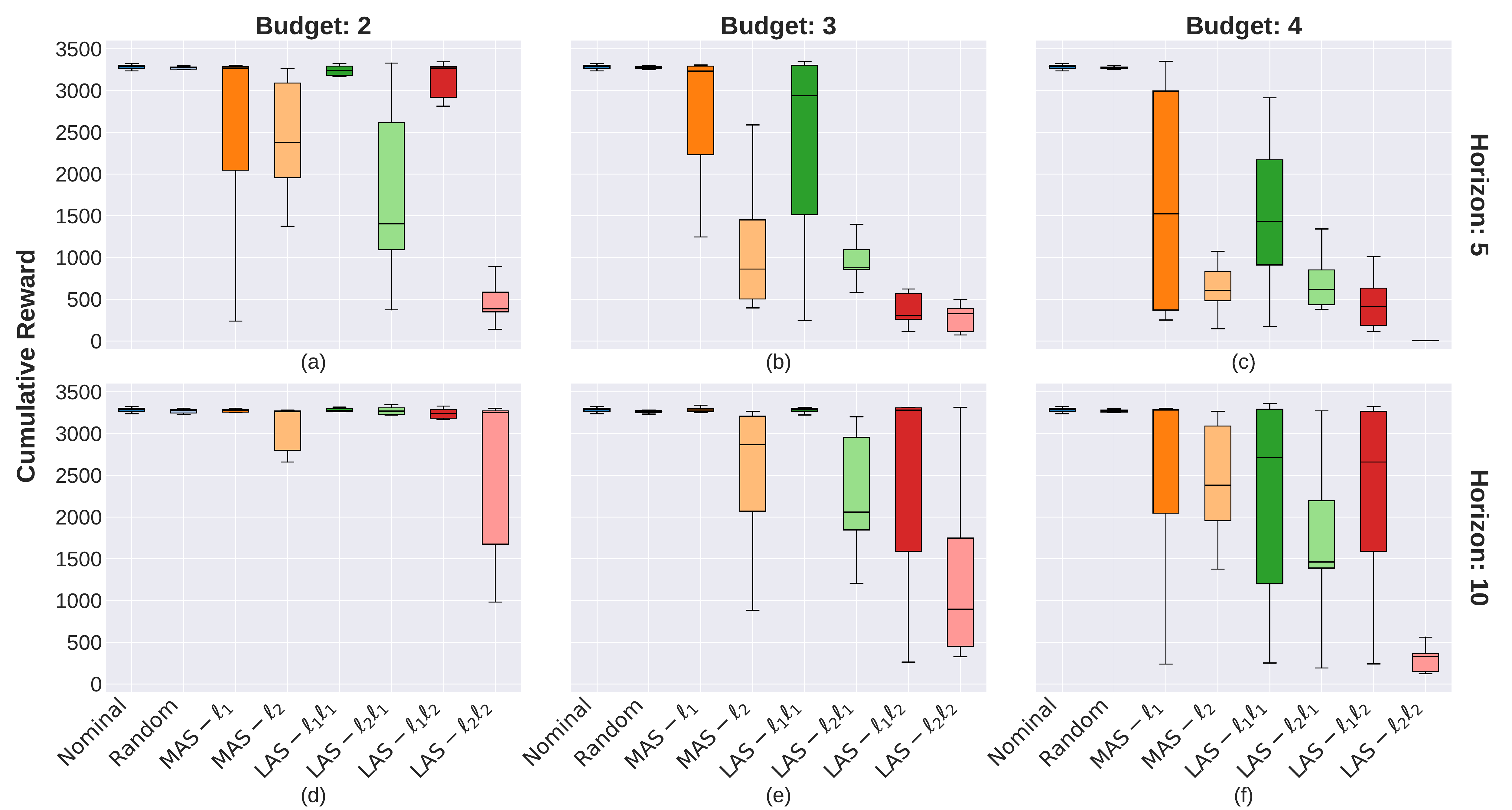}
    \caption{Box plots depicting distribution of rewards obtained by PPO agent in Mujoco Hopper environment under different attacks. In this set of experiments, we used a values of $B$ = 2, 3, and 4 for values of $H$=5 and 10. We observe similar reward trends where LAS attacks are generally stronger than MAS attacks across all values of budget and horizon.} 
    \label{fig:PPO_Hopper}
\end{figure*}

\begin{figure*}[!hb]
    \centering
    \includegraphics[width=0.85\textwidth,clip,trim={0.0in 0.0in 0.0in 0.0in}]{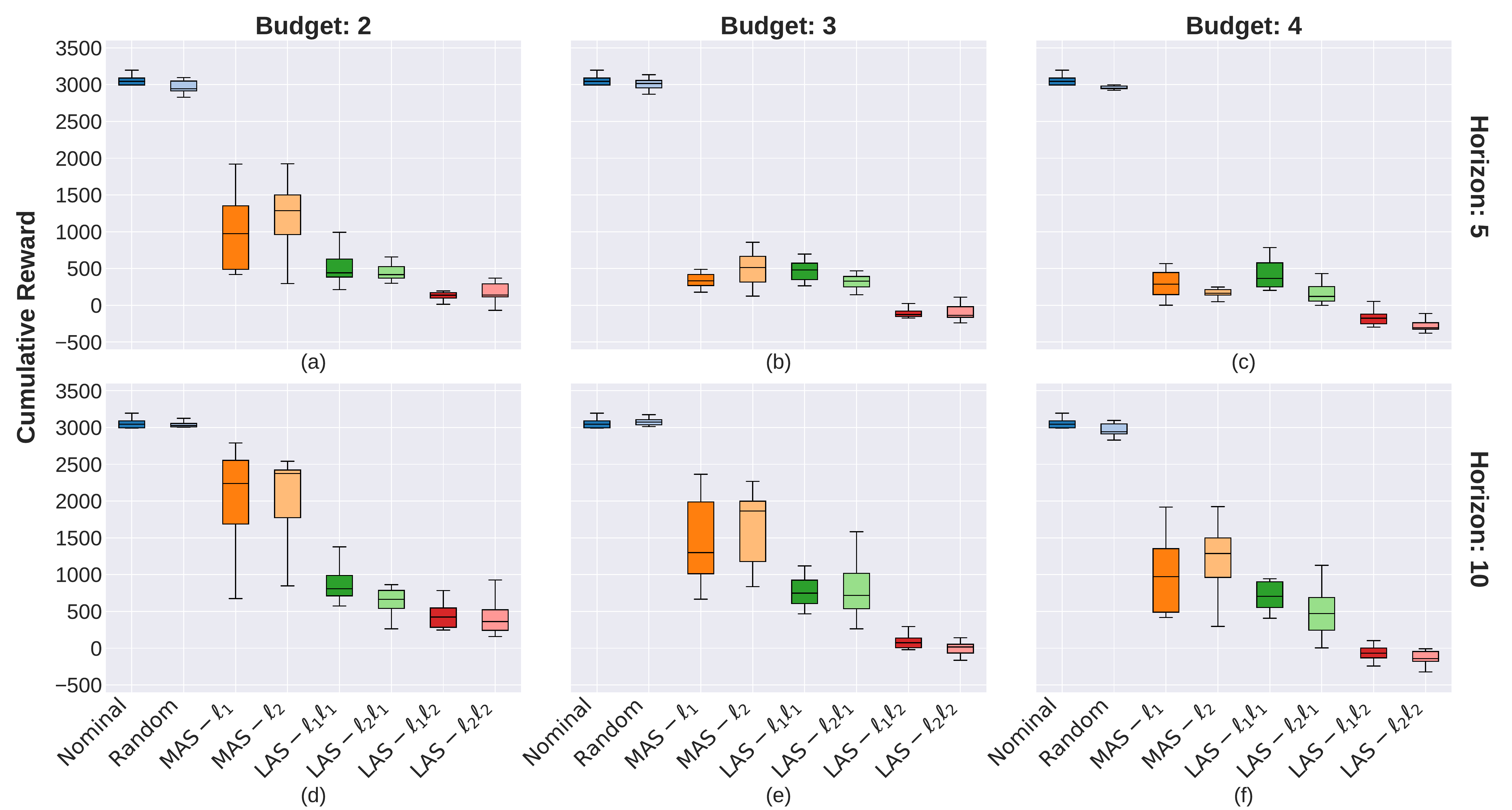}
    \caption{Box plots depicting distribution of rewards obtained by PPO agent in Mujoco Half-Cheetah environment under different attacks. In this set of experiments, we used a values of $B$ = 2, 3, and 4 for values of $H$=5 and 10. In this environment, the agent is more sensitive to the attacks for the budget values tested, where even a $B$=2 significantly decreases the distribution of rewards for MAS attacks. Nonetheless, we still observe similar reward trends where LAS attacks are generally stronger than MAS attacks across all values of budget and horizon. This suggests that the agent in this environment can easily be adversarially perturbed even with limited budget.} 
    \label{fig:PPO_Cheetah}
\end{figure*}

\begin{figure*}[ht]
    \centering
    \includegraphics[width=0.85\textwidth,clip,trim={0.0in 0.0in 0.0in 0.0in}]{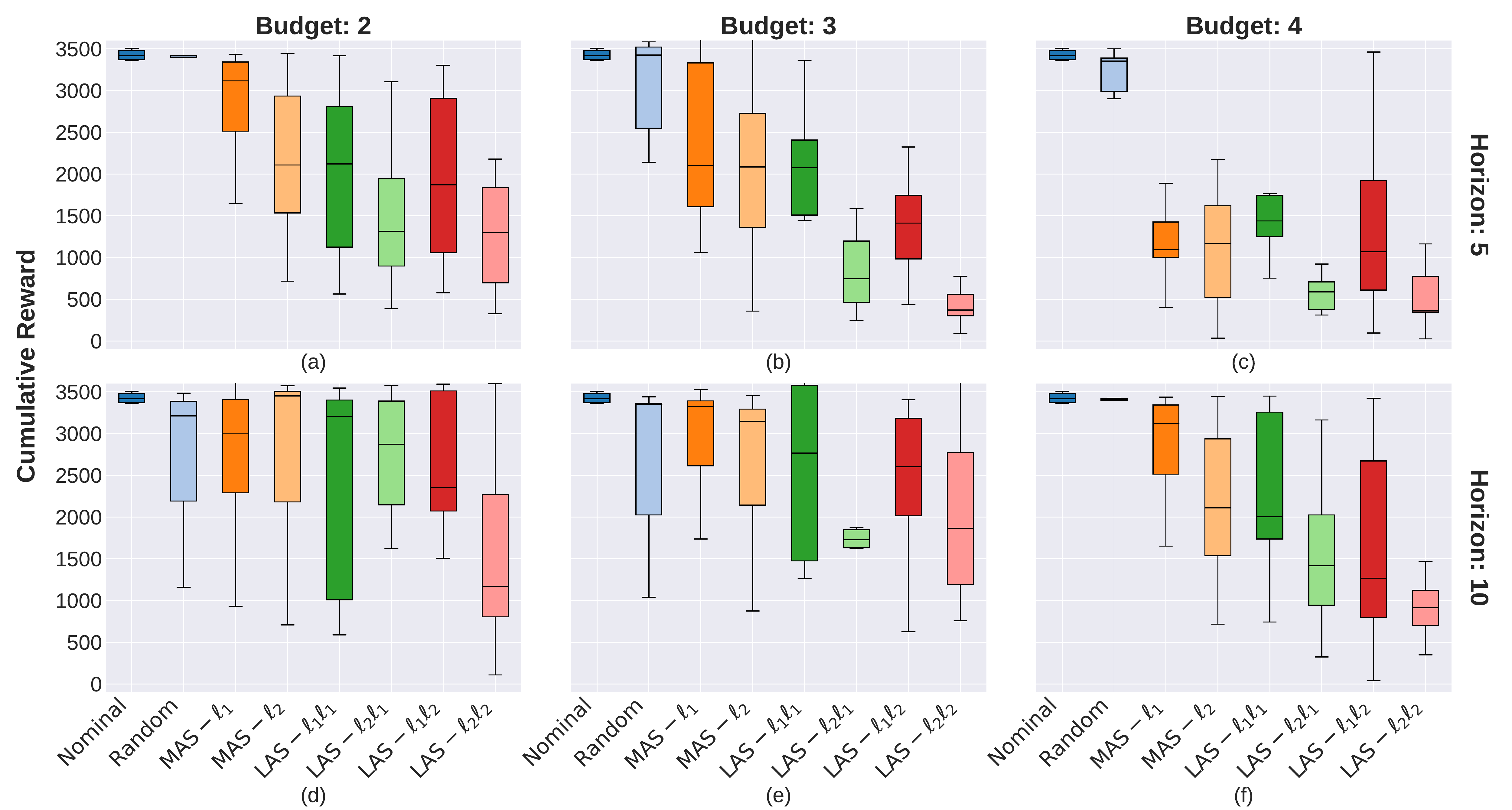}
    \caption{Box plots depicting distribution of rewards obtained by PPO agent in Mujoco Walker environment under different attacks. In this set of experiments, we used same values of $B$ = 2, 3, and 4 for values of $H$=5 and 10 as the other Mujoco environments tested above. Similar reward trends where LAS attacks decreases rewards more than MAS attacks are also observed across all values of budget and horizon. In this environment, the shifts in reward distribution are less drastic, especially for $B$=2, when comparing different attacks. This suggests that the agent might be more robust in this case and information from the environment dynamics is actually required to significantly affect the agent as seen in plot (d). } 
    \label{fig:PPO_Walker}
\end{figure*}

In addition to the attacks mounted on the agents in the 2 environments above, we also compare the effect of the attacks on a PPO agent trained in 3 different Mujoco control environments. Figures~\ref{fig:PPO_Hopper},~\ref{fig:PPO_Cheetah},~\ref{fig:PPO_Walker} illustrates the distribution of rewards obtained by the agent in the Hopper, Half-Cheetah and Walker environments respectively. In all three environments, we observe that LAS attacks are generally more effective in reducing the rewards of the agent across different values of budget and horizon, which reinforces the fact that LAS attacks are stronger than MAS attacks. However, it is also interesting to note that agents in different environments have different sensitivity to the budget of attacks. In Hopper(Figure~\ref{fig:PPO_Hopper}) and Walker(Figure~\ref{fig:PPO_Walker}), we see that MAS attacks have the effect of shifting the median and increasing the variance of the reward distortion with respect to the nominal, which highlights the fact that there are some episodes which MAS attack fails to affect the agent. In contrast, in the Half-Cheetah environment(Figure~\ref{fig:PPO_Cheetah}), we see that MAS attacks shifts the whole distribution of rewards downwards, showing that the agent is more sensitive in to MAS in this environment as compared to the other two environments. This suggests that the budget and horizon values are also hyper-parameters which should be tuned according to the environment. 
\clearpage
\newpage

\subsection{Comparison of Temporal Projections in LAS}
\begin{figure*}[h]
    \centering
    \includegraphics[width=0.8\textwidth,clip,trim={0.0in 0.0in 0.0in 0.0in}]{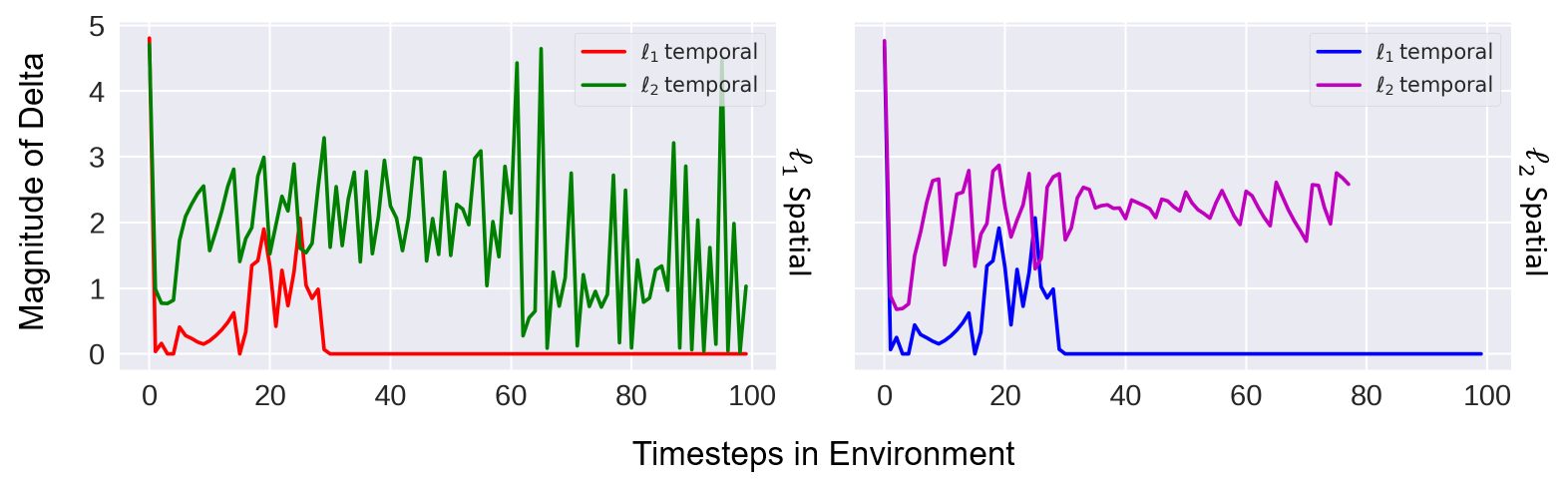}
    \caption{Comparison of $\|\delta_t \|$ used across time for a single episode in PPO Lunar Lander for different spatial projections with $\ell_1$ and $\ell_2$ temporal projection. Left plot illustrates $\ell_1$ spatial projection and right plot shows $\ell_2$ spatial projection. In both plots, the magnitude of attacks with $\ell_1$ temporal projection attacks dropped to zero from time step 30. However, the magnitude of attacks in $\ell_2$ temporal projection remains high through the episode. Hence, we observe that $\ell_1$ temporal projections essentially allows the agent sufficient time to recover from earlier attacks. In the case of Lunar Lander, the agent might prevent a severe crash while landing or recover from a horizontal thruster boost induced by the attack.}
    \label{fig:PPO_LL_delta}
\end{figure*}

\begin{figure*}[h]
    \centering
    \includegraphics[width=0.8\textwidth,clip,trim={0.0in 0.0in 0.0in 0.0in}]{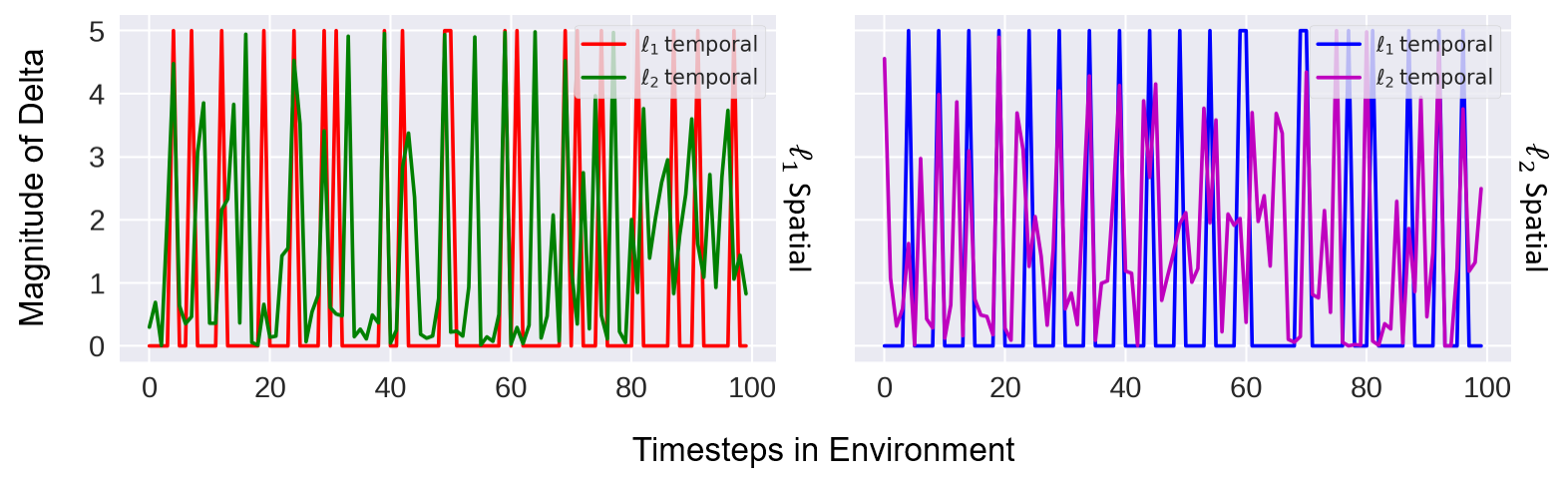}
    \caption{$\| \delta_t \|$ usage plot of DDQN agent in Lunar Lander across time for a single episode. Left subplot illustrates $\ell_1$ spatial attacks while right subplot shows $\ell_2$ spatial attacks. In each subplot, attacks with $\ell_1$ time projection attacks exhibit periodic spiked patterns while $\ell_2$ time projection attacks are constantly activated with $ \| \delta_t \|$ never reaching zero. Since $\ell_1$ time projection attacks periodically inject $\delta_t$ into nominal actions of the DDQN agent, the agent has the opportunity to recover from the attacks. }
    \label{fig:DDQN_LL_delta}
\end{figure*}

\begin{figure*}[h]
    \centering
    \includegraphics[width=0.8\textwidth,clip,trim={0.0in 0.0in 0.0in 0.0in}]{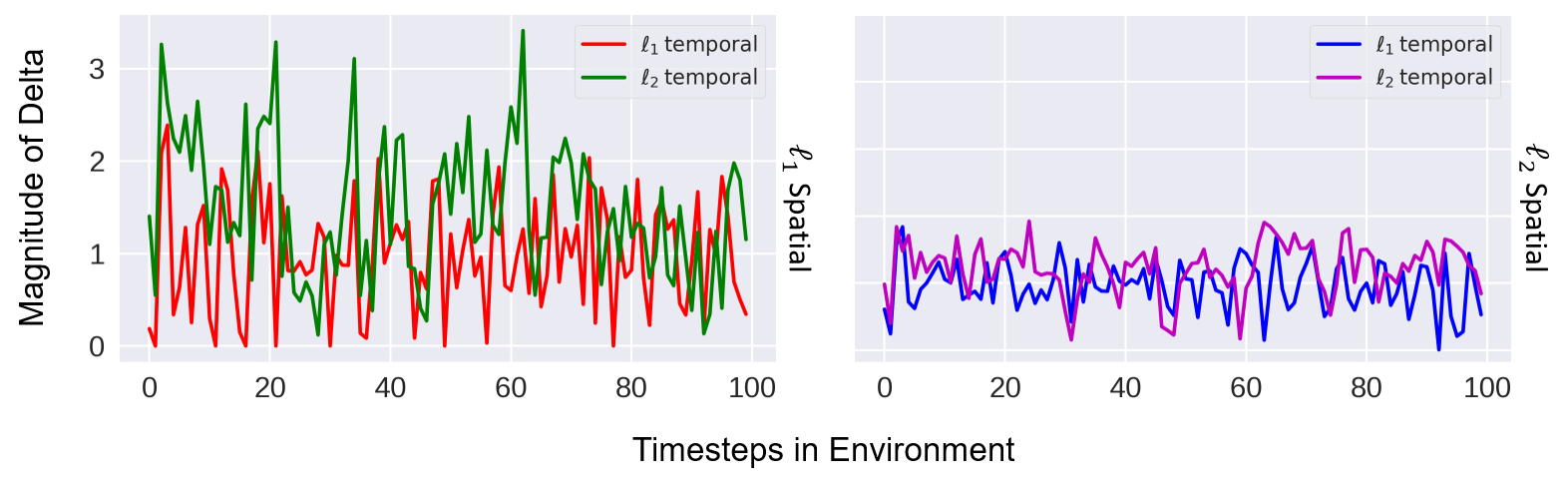}
    \caption{ $\| \delta_t \| $ usage plot for PPO agent in Bipedal-Walker across time for a single episode. Left subplot illustrates $\ell_1$ spatial attacks while right subplot shows $\ell_2$ spatial attacks. In this figure, both $\ell_1$ and $ell_2$ are seemingly well distributed, although the magnitude of $\| \delta_t \|$ used for both projection schemes are evidently lesser than the other agents in other environments. We speculate that this is due to the nature of the policy learnt by the PPO agent. In this environment, the PPO agent has learnt a policy to operate the Bipedal Walker by bending a knee joint and using the other knee joint to drag itself forward. Hence, in this situation, the agent has learnt a strong stable walking gait and there is not much room for $\delta_t$ to be applied. }
    \label{fig:PPO_BW_delta}
\end{figure*}

\begin{figure*}[h]
    \centering
    \includegraphics[width=0.8\textwidth,clip,trim={0.0in 0.0in 0.0in 0.0in}]{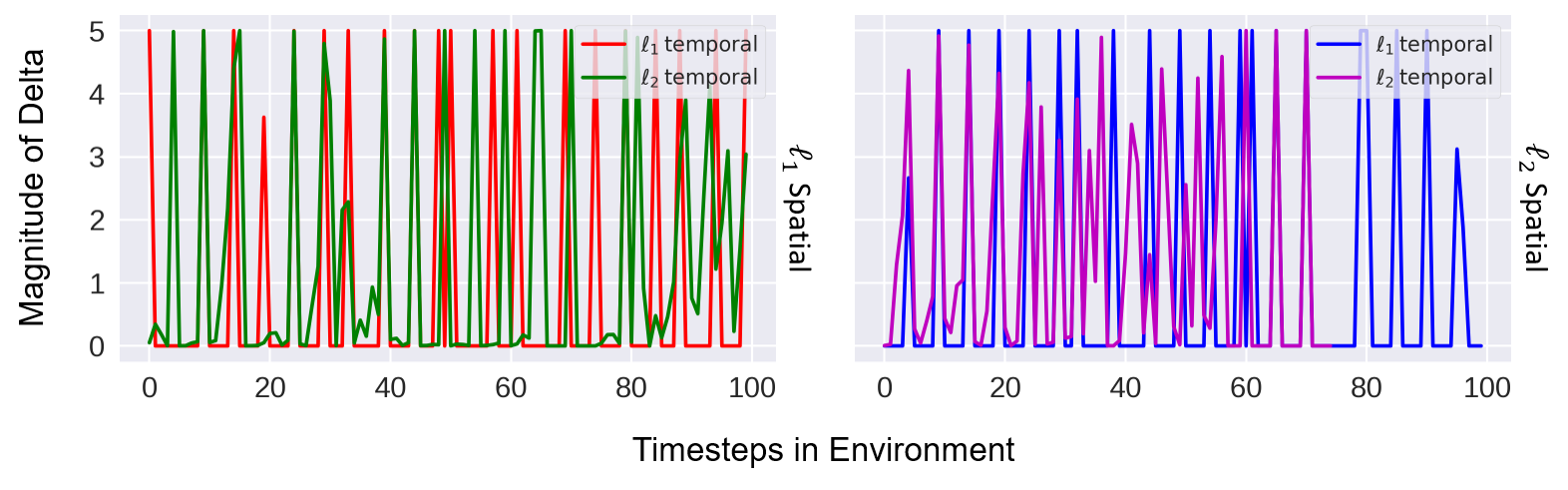}
    \caption{$\| \delta_t \|$ usage plot for DDQN agent in Bipedal-Walker across time for a single episode. Left subplot illustrates $\ell_1$ spatial attacks while the right subplot shows $\ell_2$ spatial attacks. A trend similar to Fig~\ref{fig:DDQN_LL_delta} is observed here where the $\| \delta_t \|$ are utilized in sparse and concentrated instances for $\ell_1$ temporal projections in compared to $\ell_2$ temporal projections.}
    \label{fig:DDQN_BW_delta}
\end{figure*}

\vspace{-100 pt}
In this section, we present additional visualizations to further understand why $\ell_2$ temporal projections results in more severe attacks in compared to $\ell_1$ temporal projections. Figure~\ref{fig:PPO_LL_delta}, ~\ref{fig:DDQN_LL_delta}, ~\ref{fig:PPO_BW_delta} and~\ref{fig:DDQN_BW_delta} presents the $\| \delta_t \| $ usage plot across 100 time steps for both PPO and DDQN in the Lunar Lander and Bipedal-Walker environment. The left subplot represents $\ell_1$ projections in the spatial dimension while the right subplot represents $\ell_2$ projections in the spatial dimensions. These plots directly compare the difference in amount of $\| \delta_t \| $ used between $\ell_1$ and $\ell_2$ temporal projections for both $\ell_1$ and $\ell_2$ spatial attacks. \\
\\
In most cases with the exception of Figure~\ref{fig:PPO_BW_delta}, we see a clear trend that $\ell_1$ temporal projections results in a sparser but more concentrated peaks of $\| \delta \| $ utilization (corresponding to a few instance of strong attacks). In contrast, $\ell_2$ temporal projections results in a more distributed but frequent form of of $\| \delta \| $ utilization (corresponding to more frequent but weak instances of attacks). We note that while $\ell_1$ projections produces stronger attacks, there is a diminishing return on allocating more attacks to a certain time point as after a certain limit. Hence, this explains the weaker effect of $\ell_1$ temporal projections since it concentrates the attacks to a few points but ultimately gives time for the agent to recover. In contrast, $\ell_2$ temporal projections distributes the attacks more frequently that causes the agent to follow a diverging trajectory that is hard to recover from.\\
\\
As an anecdotal example in the Lunar Lander environment, we observe that attacks with $\ell_1$ temporal projection tend to turn off the vertical thrusters of the lunar lander. However, due to the sparsity of the attacks, the RL agent could possibly be fire the upward thrusters in time to prevent a free-fall landing. With $\ell_2$ temporal projections, the agent is attacked continuously. Consequently, the agent has no chance to return to a nominal state and quickly diverges towards a terminal state.  
\vspace{-100pt}
\subsection{Ablation Study}
\begin{figure*}[h]
    \centering
    \includegraphics[width=0.8\textwidth,clip,trim={0.0in 0.0in 0.0in 0.0in}]{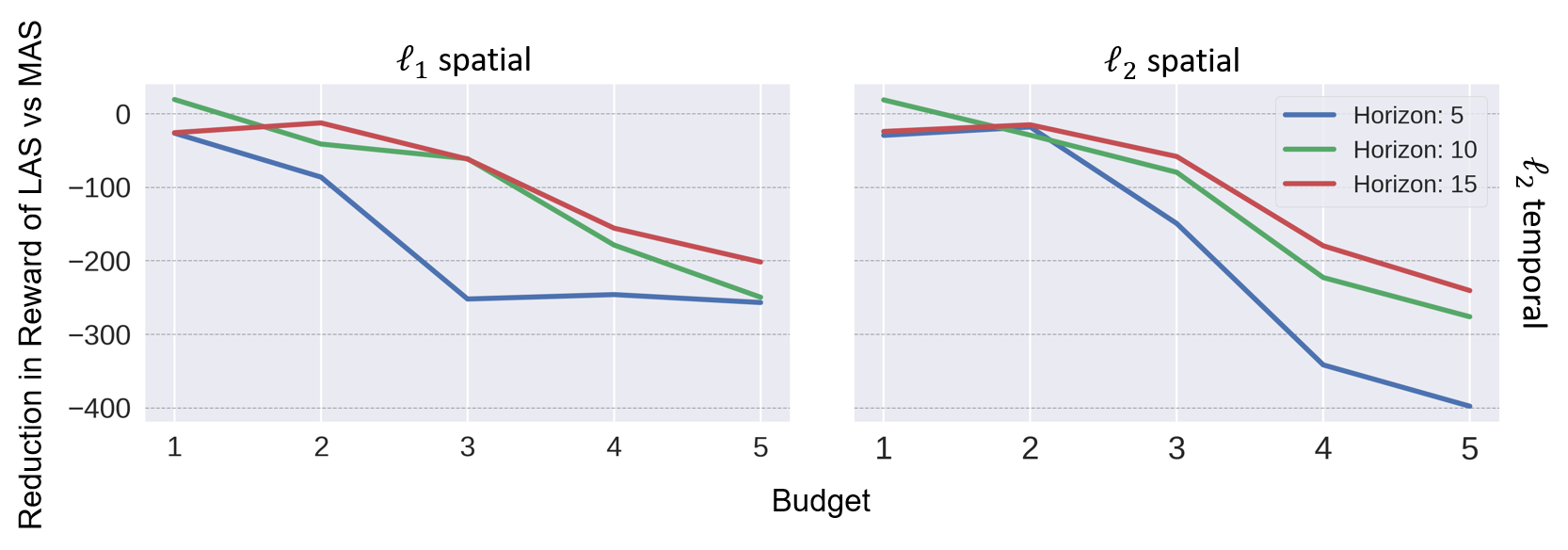}
    \caption{Ablation study for PPO Lunar Lander showing effectiveness of attacks comparing LAS with MAS for $\ell_2$ time projection attacks. Left and right subplot shows $\ell_1$ and $\ell_2$ spatial projection respectively. Each subplot contains different lines representing different horizon, where budget is incrementally increased along each horizon. Both $\ell_1$ and $\ell_2$ spatial projection scales monotonically with increasing budget.}
    \label{fig:PPO_LL_ablation}
\end{figure*}

\begin{figure*}[h]
    \centering
    \includegraphics[width=0.8\textwidth,clip,trim={0.0in 0.0in 0.0in 0.0in}]{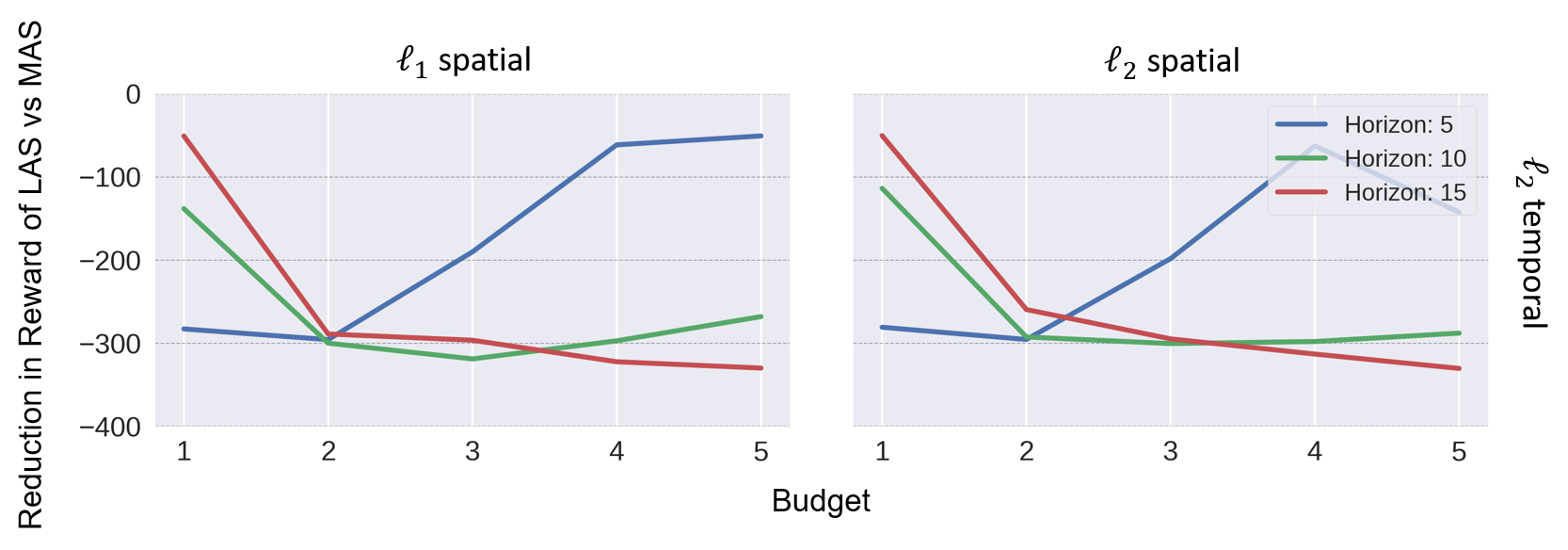}
    \caption{Ablation study for DDQN Lunar Lander showing effectiveness of attacks comparing LAS with MAS. Left subplot shows $\ell_1$ spatial attacks while right subplot shows $\ell_2$ spatial attacks. Both plots are in $\ell_2$ temporal projection attacks. Each subplot contains different lines representing different horizon, where budget is incrementally increased along each horizon. }
    \label{fig:DDQN_LL_ablation}
\end{figure*}

\begin{figure*}[ht]
    \centering
    \includegraphics[width=0.8\textwidth,clip,trim={0.0in 0.0in 0.0in 0.0in}]{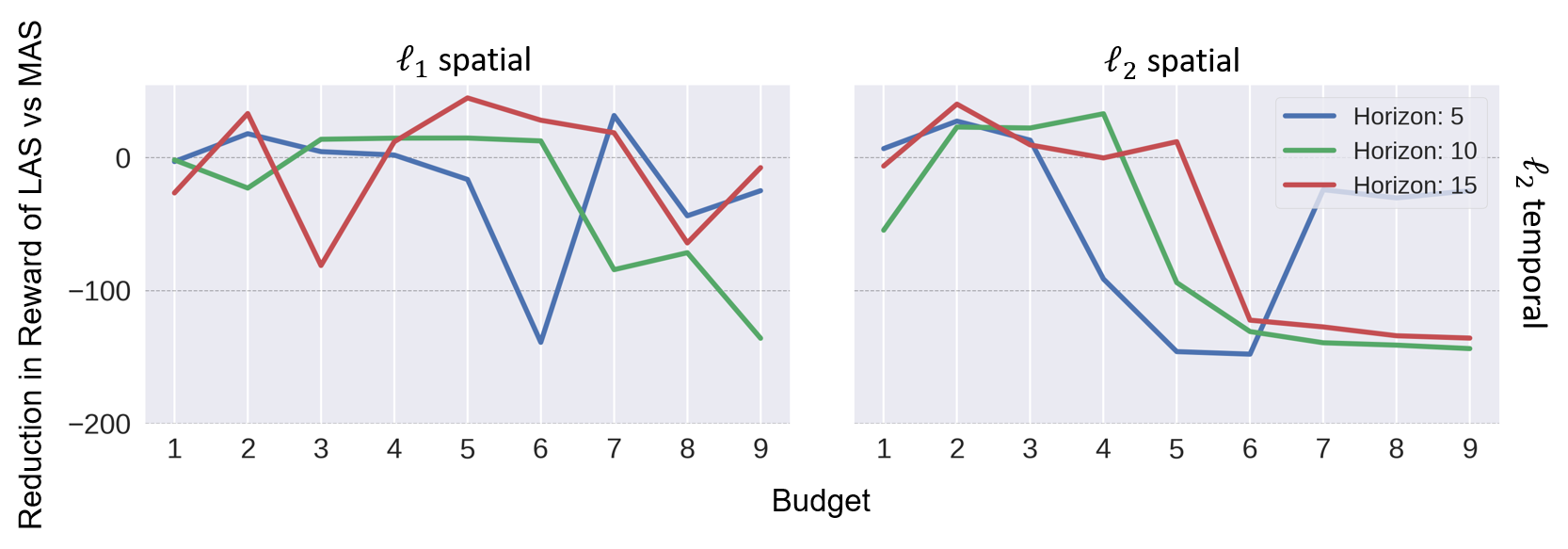}
    \caption{Ablation study for PPO Bipedal-Walker showing effectiveness of attacks comparing LAS with MAS. Left subplot shows $\ell_1$ spatial attacks while right subplot shows $\ell_2$ spatial attacks. Both plots are in $\ell_2$ temporal projection attacks. Each subplot contains different lines representing different horizon, where budget is incrementally increased along each horizon. }
    \label{fig:PPO_BW_ablation}
\end{figure*}

\begin{figure*}[h!]
    \centering
    \includegraphics[width=0.8\textwidth,clip,trim={0.0in 0.0in 0.0in 0.0in}]{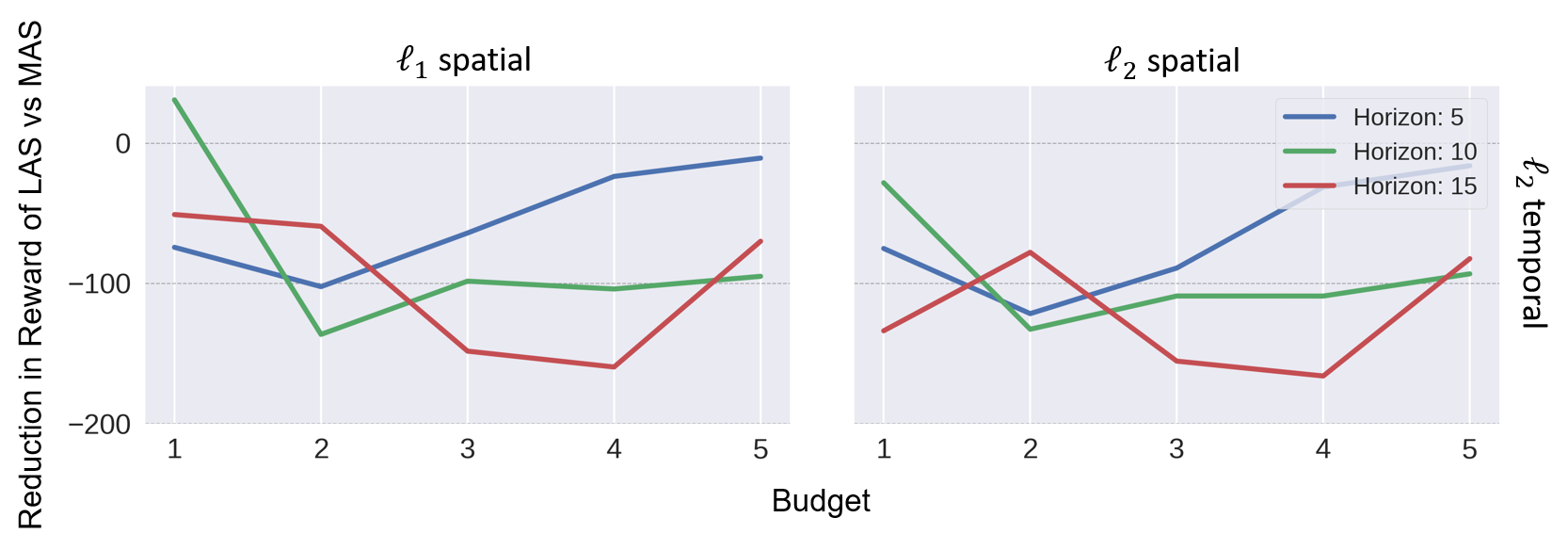}
    \caption{Ablation study for DDQN Bipedal-Walker showing effectiveness of attacks comparing LAS with MAS. Left subplot shows $\ell_1$ spatial attacks while right subplot shows $\ell_2$ spatial attacks. Both plots are in $\ell_2$ temporal projection attacks. Each subplot contains different lines representing different horizon, where budget is incrementally increased along each horizon. }
    \label{fig:DDQN_BW_ablation}
\end{figure*}

\vspace{-95pt}
For this section, we present an ablation study to investigate the effect of different budget and horizon parameters on the effectiveness of LAS vs MAS. As mentioned in the main manuscript, we take the difference of each attack's reduction in rewards (i.e. attack - nominal) and visualize how much rewards LAS reduces as compared to MAS under different conditions of $B$ and $H$. Fig~\ref{fig:PPO_LL_ablation} illustrates the ablation study of a PPO agent in Lunar Lander. The figure is categorized by different spatial projections, where $\ell_1$ spatial projections are shown on the left figure while $\ell_2$ spatial projections are shown on the right. Both subplots are shown for $\ell_2$ time projection attacks. Each individual subplot shows three different lines with different $H$, with each line visualizing the change in mean cumulative reward as budget increases along the x-axis. As budget increases, attacks in both $\ell_1$ and $\ell_2$ spatial projection shows a monotonic decrease in cumulative rewards. Attacks in each spatial projection with a $H$ value of 5 shows different trends, where $\ell_2$ decreases linearly with increasing budget while $\ell_1$ became stagnant after $B$ value of 3. This can be attributed to the fact that the attacks are more sparsely distributed in $\ell_1$ attacks, causing most of the perturbations to be distributed into one action dimension. Thus, as budget increases, we see a diminishing return of LAS attacks since attacking a single action dimension beyond a certain limit doesn't decrease reward any further.
The study was also conducted for PPO Bipedal-Walker and both DDQN Lunar Lander and Bipedal-Walker as shown in Figure~\ref{fig:PPO_BW_ablation}, ~\ref{fig:DDQN_LL_ablation} and ~\ref{fig:DDQN_BW_ablation} respectively. We only consider attacks in $\ell_2$ temporal projection attacks for both $\ell_1$ and $\ell_2$ spatial projections. At a glance, we see different trends across each figures due to the different environment dynamics. However, in all cases, the decrease in reduction of rewards is always lesser than or equals to zero, which infers that LAS attacks are at least as effective than MAS attacks. We observed that attacks for horizon value of 5 becomes ineffective after a certain budget value. This shows that after some budget value, MAS attacks are as effective as LAS attacks because LAS might be operating at maximum attack capacity. Interesting to note that Bipedal-Walker for PPO needed a higher budget compared to the DDQN counterpart due to the PPO being more robust to attacks. 

\subsection{Effect of Horizon Parameter in LAS} 
In this section, we further describe the effect of horizon parameter $H$ on the effectiveness of LAS attacks that we empirically observed. $H$ defines a fixed time horizon (e.g., steps in DRL environments) to expend a given budget $B$. For a fixed $B$ and short $H$, LAS favors injecting stronger perturbations in each step. Hence, we would intuitively hypothesize that given a shorter $H$, the severity of LAS attacks will increase as $H$ decrease, as shown by the reduction in rewards between MAS and LAS in Figure 3 of the main paper. In most cases, the reduction is negative, hence showing that LAS attacks are indeed more severe. However in some cases as shown in Figure 8, 9 \& 10, a shorter $H$ does result in LAS being not as effective as a longer $H$ (though still stronger than MAS as evident from negative values of y-axis). This can be attributed to the nonlinear reward function of the environments and consequent failure modes of the agent. For example, attacks on Lunar Lander PPO agent causes failure by constantly firing thruster engines to prevent Lunar Lander from landing, hence accumulating negative rewards over a long time. In contrast, attacks on the DQN agent causes Lunar Lander to crash immediately, hence terminating the episode before too much negative reward is accumulated. Thus, while the effect of $H$ on LAS attacks sometimes do not show a consistent trend, we it is a key parameter that can be tuned to control the failure modes of the RL agent. 

\subsection{Action Space Dimension Decomposition}

\begin{figure*}[!h]
    \centering
    \includegraphics[width=0.95\columnwidth,clip,trim={3.50in 1.8in 3.5in 1.0in}]{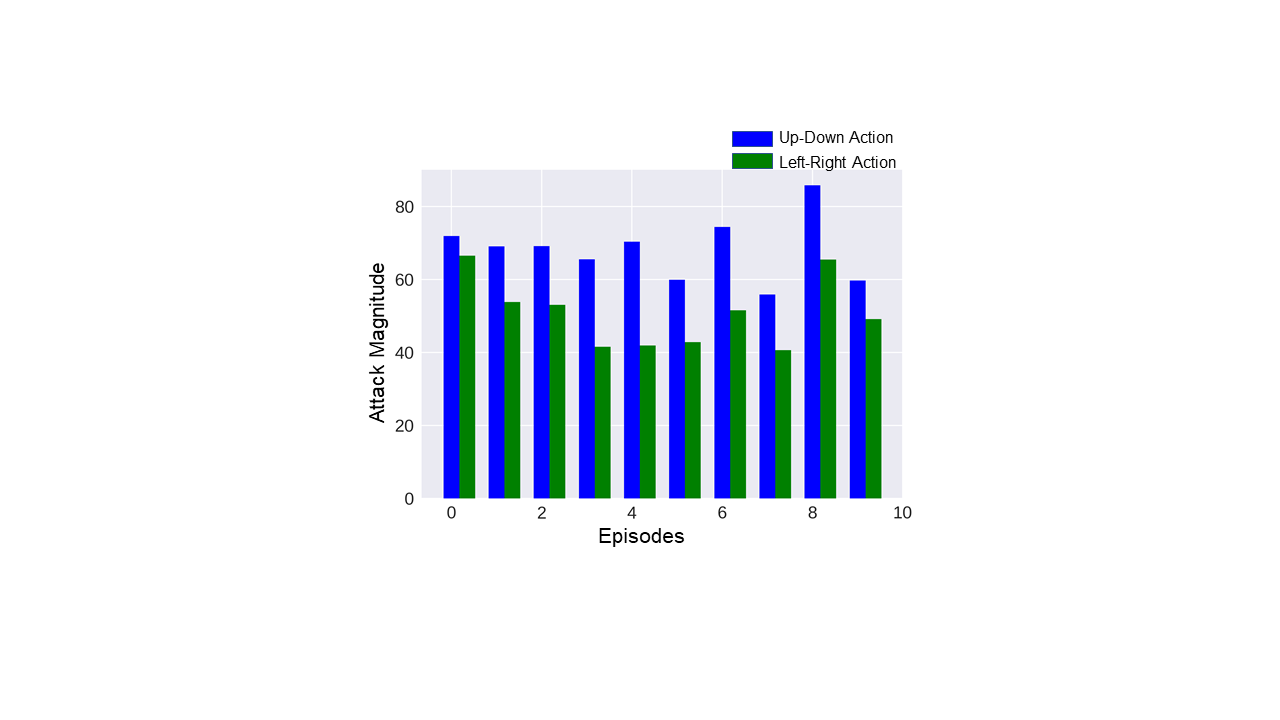}
    \caption{Magnitude of attack with respect to different episodes for Lunar Lander environment with DoubleDQN RL agent. The two colors (Blue and Green) in the bar plots represent the attacks allocated to the two action space dimensions, Up-Down action and Left-Right action, respectively. The attack schema is LAS attack with a budget of 4 and Horizon of 5.}
    \label{fig:DDQN_LL_dimension}
\end{figure*}

\begin{figure*}[!t]
    \centering
    \includegraphics[width=0.95\columnwidth,clip,trim={3.50in 1.8in 3.5in 1.0in}]{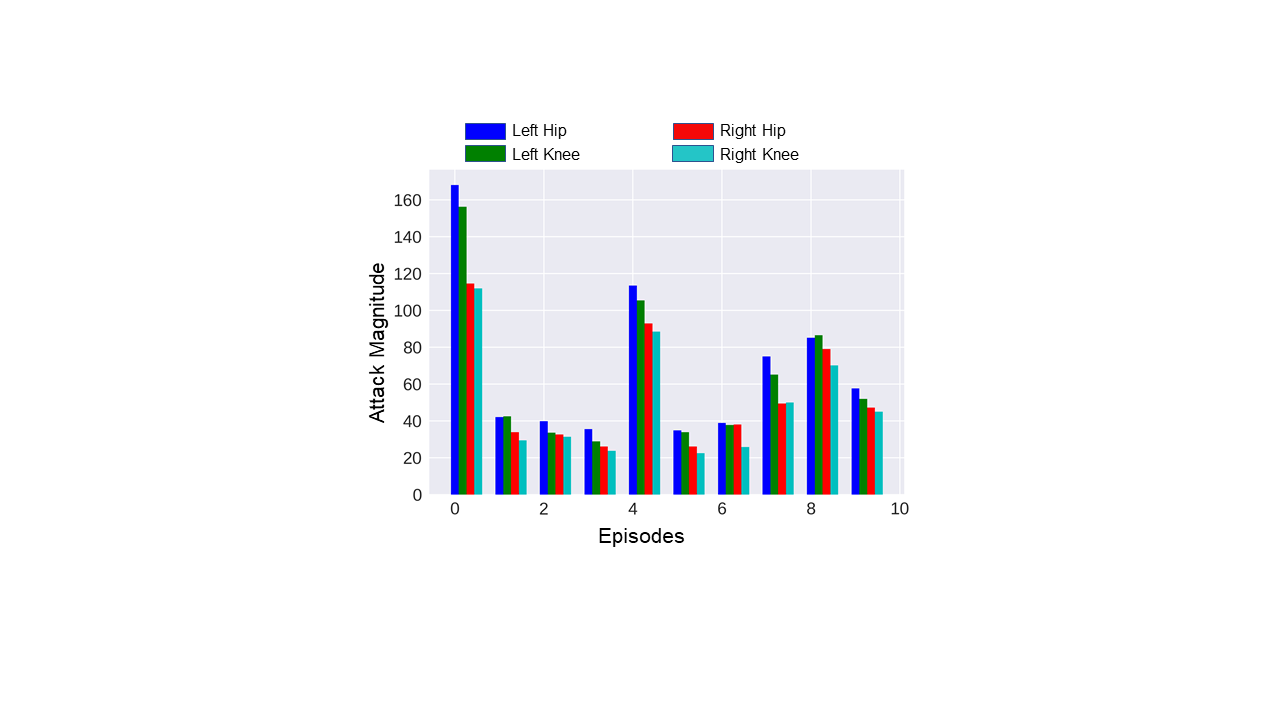}
    \caption{Magnitude of attack with respect to different episodes for Bipedal Walker environment with DoubleDQN RL agent. The four colors (Blue, Green, Red and Cyan) in the bar plots represent the attacks allocated to the four action space dimensions, Left Hip, Left Knee, Right Hip and Right Knee actions, respectively. The attack schema is LAS attack with a budget of 4 and Horizon of 5.}
    \label{fig:DDQN_BW_dimension}
\end{figure*}

We provide additional results on using the LAS attack scheme as a tool to understand the RL agent's action space vulnerabilities for a DoubleDQN agent in both Lunar Lander and Bipedal Walker environment. It is interesting to note in Figure~\ref{fig:DDQN_LL_dimension}, even with agents trained with a different philosophy (value-based vs policy based, shown in main manuscript), the attack scheme still distributes the attack to a similar dimension (Up-Down action for Lunar Lander), which highlights the importance of the that particular dimension. In Figure \ref{fig:DDQN_BW_dimension}, we show the outcome of LAS attack scheme on Bipedal Walker environment having four action space dimensions. The four joints of the bipedal walker, namely Left Hip, Left Knee, Right Hip and Right Knee are attacked in this case, and from Figure \ref{fig:DDQN_BW_dimension}, we see that the left hip is attacked more than any other action dimension in most of the episodes. This supports our inference that LAS attacks can bring out the vulnerabilities in the action space dimensions (actuators in case of CPS RL agents).

\clearpage
\newpage

\vspace{-80pt}

\end{document}